\documentclass[fleqn,twocolumn,10pt]{wlscirep}
\usepackage[utf8]{inputenc}
\usepackage{subfigure}
\usepackage[T1]{fontenc}
 \usepackage{lineno}
 
\usepackage[english]{babel}
\usepackage{colortbl}
\usepackage{newfloat}
\definecolor{lightpastelpurple}{rgb}{0.7,0.55,0.43}
\definecolor{LightCyan}{rgb}{0.88,1,1}
\definecolor{lightgreen}{rgb}{0.82,0.94,0.75}
\DeclareFloatingEnvironment[name={Supplementary Figure},fileext=lsf,listname={List of Supplementary Figures}]{suppfigure}
\title{Building Trust in Virtual Immunohistochemistry: Automated Assessment of Image Quality}

\author[1]{Tushar Kataria}
\author[1,+]{Shikha Dubey}
\author[3]{Mary Bronner}
\author[3,4]{Jolanta Jedrzkiewicz}
\author[2]{Ben J. Brintz}
\author[1*]{Shireen Y. Elhabian}
\author[3*]{Beatrice S. Knudsen}
\affil[1]{Scientific Computing and Imaging Institute \& Kahlert School of Computing, University of Utah, Salt Lake City-84112, USA}
\affil[2]{Department of Internal Medicine, Division of Epidemiology, University of Utah, Salt Lake City-84112, USA}
\affil[3] {Department of Pathology, University of Utah, Salt Lake City-84112, USA}
\affil[4] {ARUP Laboratories, Salt Lake City, 84112,USA}
\affil[+]{work done while at University of Utah}
\affil[*]{Co-senior authors}

\begin{abstract}
Deep learning models can generate virtual immunohistochemistry (IHC) stains from hematoxylin and eosin (H\&E) images, offering a scalable and low-cost alternative to laboratory IHC. However, reliable evaluation of image quality remains a challenge as current texture- and distribution-based metrics quantify image fidelity rather than the accuracy of IHC staining. Here, we introduce an automated and accuracy grounded framework to determine image quality across sixteen paired or unpaired image translation models. Using color deconvolution, we generate masks of pixels stained brown (i.e., IHC-positive) as predicted by each virtual IHC model. We use the segmented masks of  real and virtual IHC to  compute stain accuracy metrics (Dice, IoU, Hausdorff distance) that directly quantify correct pixel - level labeling without needing expert manual annotations. Our results demonstrate that conventional image fidelity metrics, including Fréchet Inception Distance (FID), peak signal-to-noise ratio (PSNR), and structural similarity (SSIM), correlate poorly with stain accuracy and pathologist assessment. Paired models such as PyramidPix2Pix and AdaptiveNCE achieve the highest stain accuracy, whereas unpaired diffusion- and GAN-based models are less reliable in providing accurate IHC positive pixel labels. Moreover, whole-slide images (WSI) reveal performance declines that are invisible in patch-based evaluations, emphasizing the need for WSI-level benchmarks. Together, this framework defines a reproducible approach for assessing the quality of virtual IHC models, a critical step to accelerate translation towards routine use by pathologists.
\end{abstract}
\begin{document}
\flushbottom
\maketitle
%
%

\section*{Introduction}

Image-to-image (I2I) translation in computational pathology enables the generation of diverse virtual histopathology stains. These include virtual H\&E stains \cite{ho2024disc,ozyoruk2021deep}, chemical stains highlighting features such as fibrosis \cite{levy2021large}, basement membranes \cite{yang2022virtual,gadermayr2018way}, amyloidosis \cite{yang2024virtual}, or lipid deposits \cite{wieslander2021learning,li2024label}, as well as virtual immunohistochemistry \cite{li2023adaptive,dubey2023structural,kataria2024staindiffuser} and immunofluorescence \cite{bian2024hemit,wu2025rosie} for discerning cell types and differentiation states. These models operate in the same way as popular text-conditional generative models such as DALL·E \cite{ramesh2021zero} and ChatGPT \cite{achiam2023gpt}, but instead of generating images from text prompts, they generate images conditioned on a reference input image \cite{peng2024controlnext,zhang2023adding,zhu2017unpaired,isola2017image}.
Before deploying virtual stains in the clinic they need to be carefully evaluated.  Comprehensive, multiparametric evaluation frameworks  must assess \textit{visual fidelity}, ensuring accurate reconstruction of cell morphology and tissue architecture, and \textit{staining accuracy}, confirming the correct identification of stain-positive cells. Only through rigorous evaluation along these axes can virtual staining approaches establish reliability and trustworthiness for clinical adoption.

State-of-the-art medical image-to-image translation methods \cite{brodsky2025generative,winter2025utilizing,bai2023deep,li2023adaptive,kataria2024staindiffuser,liu2022bci,kataria2025implicitstainer,huijben2024generating,livieris2024evaluation,zamzmi2024scorecards,dubey2023structural,dubey2024vims} are typically evaluated using texture-based metrics (Signal to noise ratio: PSNR, Structural Similarity: SSIM, Mean Square Error: MSE) and Distribution-based metrics such as Fréchet Inception Distance (FID) \cite{heusel2017gans}, Kernel Inception Distance (KID) \cite{binkowski2018demystifying}, and distribution precision and recall \cite{kynkaanniemi2019improved} quantify the diversity and coverage of high-dimensional encoded features, i.e., the latent-space representations extracted from a foundational or pretrained model. However, both families of metrics assess only visual fidelity and not the staining accuracy of virtual immunohistochemistry (IHC) images \cite{dubey2023structural,kataria2024staindiffuser}.  Distribution-based measures rely on population-level latent space statistics (mean and covariance matrices) and cannot capture staining accuracy at the pixel level \cite{dubey2023structural,zamzmi2024scorecards,livieris2024evaluation}. Texture-based metrics quantify pixel-level differences between generated and real images, measuring perceptual or structural similarity and not the spatially resolved expression of specific proteins \cite{liu2022bci,kataria2024staindiffuser}. They can be used to evaluate the quality of tissue architecture and cell morphology, but not the correctness of IHC positive pixel labels. In addition, the assessment of image quality by distribution and texture metrics differs from that of human evaluators who are insensitive to small structural variations and instead prioritize the accuracy of information conveyed by color of cells and tissues \cite{Gonzlez2008DigitalIP,fei2012perceptual,breger2025study,yang2015perceptual,zhai2020perceptual}.

Given the limitations of existing metrics, the evaluation by a domain expert pathologist remains the standard for assessing the quality of computer-generated (or virtual) images \cite{rivenson2019phasestain,rivenson2019virtual,ho2025f2fldm,ho2024disc}. However, domain expert reviews are costly, time‑consuming, and limited in scale, comprising a few hundred to a thousand samples.  This makes virtual whole-slide image evaluation infeasible. In addition, pathologists may not be able to determine which cells in the image are correctly stained unless the stain is associated with characteristic morphologic features of cell type and/or cell state. Therefore, there is a need for automated assessment of stain accuracy at scale. A first step towards this goal is to determine whether widely used distribution- and texture- based metrics \cite{liu2022bci,li2023adaptive,kang2021stainnet,jaume2024multistain,kataria2024staindiffuser,kataria2025implicitstainer,dubey2023structural,ho2025f2fldm} reliably capture both visual fidelity and staining accuracy in medical image translation tasks \cite{huijben2024generating,raggio2025fedsynthct,li2023ct,10081095}.

\begin{figure*}[!htb]
    \centering
    \includegraphics[width=1.0\linewidth]{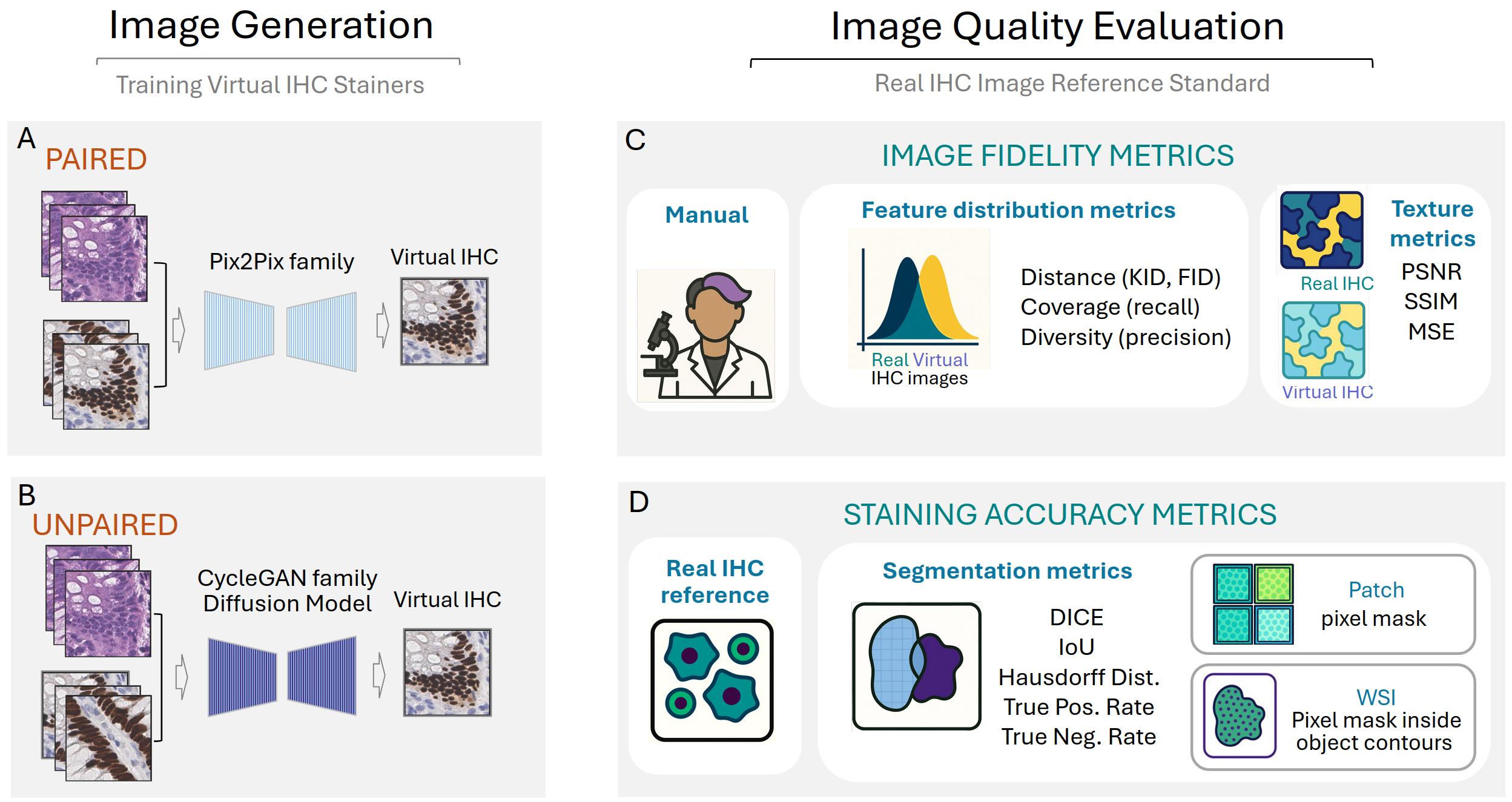}
    \caption{\textbf{Workflow to generate virtual IHC images and evaluate their quality.} \textbf{A.} Paired H\&E and IHC tiles extracted from the exact same tissue stained with H\&E and restained with IHC are used to train Pix2Pix family models to generate virtual IHC images. \textbf{B.} Unpaired H\&E and IHC tiles from different tissues stained with H\&E and IHC are used to train cycle-GAN family or diffusion models. \textbf{C.} Evaluation of image quality utilizes standard image fidelity metrics, including manual, distribution-based and texture-based metrics. \textbf{D.} Stain accuracy metrics consist of segmentation metrics to determine if the correct pixels are colored in computer generated stains. Stain accuracy is determined on both image tiles and whole slide images (WSI). FID - Frechet Inception Distance, KID - Kernel Inception Distance, PSNR - Peak Signal-to-Noise Ratio, SSIM – Structural Similarity Index, MSE- Mean Square Error, DICE – DICE Similarity Coefficient, IoU – Intersection over Union.}
    \label{fig:graphical_abstract}
\end{figure*}

In this work, we introduce a comprehensive evaluation framework that in addition to traditional distribution and texture metrics proposes metrics to assess staining accuracy. We deconvolute \cite{landini2021colour,bianconi2020experimental,ruifrok2001quantification} IHC images into hematoxylin and DAB, a brown chromogen, and compare DAB masks between real and virtual images. This approach allows to automatically identify IHC-positive pixels, enabling scalable evaluation of staining accuracy without the need of manual annotations. A critical prerequisite for developing automated staining accuracy metrics is access to pixel-accurate ground-truth of the IHC stain, which can be generated by H\&E and IHC re-staining of the same tissue \cite{kataria2023automating,bian2024hemit,rivenson2019phasestain}.  To demonstrate and validate that our findings are generalizable and not biased toward particular I2I architectures or model designs, we assess image quality metrics and their correlations across sixteen diverse image translation models, including both GAN and diffusion frameworks. Our findings reveal that texture- and distribution-based metrics correlate poorly with staining accuracy, supporting the premise that image fidelity and stain accuracy capture different aspects of image quality. We also highlight challenges in evaluating whole-slide images (WSIs) compared to the standard tile-based assessments. Together, these results underscore the need for broader, scalable, and multifaceted evaluation frameworks for virtual staining and, more broadly, for medical image translation to enhance the trust of pathologists and accelerate clinical adoption.

\section*{Results}\label{sec:Results}

\paragraph{Computer-generated virtual IHC images and, their quality assessments.} 

Image-to-image (I2I) translation uses AI or deep learning models to generate IHC stains from digital H\&E images, i.e., virtual staining. I2I models for virtual staining fall into two main categories: (a) \textit{paired translation} (\textbf{Figure \ref{fig:graphical_abstract}A}), where H\&E and corresponding real IHC are pixel-aligned, enabling direct pixel-level supervision during training, and (b) \textit{unpaired translation }(\textbf{Figure \ref{fig:graphical_abstract}B}), where images lack direct alignment (e.g., neighboring tissue sections, tissues from different patients, or non-overlapping regions), making pixel-level correspondence impossible. Different model families are tailored to these settings. Models in the \textit{Pix2Pix family}\cite{isola2017image,liu2022bci,li2023bbdm,li2023adaptive} learn pixel-wise reconstructions from aligned H\&E–IHC pairs to achieve highly accurate mapping, although scalability is limited by the need for precise registration \cite{liu2022bci,dubey2023structural}. Models in the \textit{CycleGAN family }\cite{zhu2017unpaired,park2020contrastive,tang2021attentiongan,xie2023unpaired,xieunsupervised,torbunov2023uvcgan,wu2024stegogan,hu2022qs,kim2023unpaired,liu2017unsupervised,chen2022eccv} enable unpaired translation, allowing training on large archival cohorts but potentially introducing structural inconsistencies or hallucinations due to the absence of pixel-level correspondence loss \cite{dubey2023structural,wu2024stegogan}. Unpaired diffusion-based models \cite{kim2023unpaired} offer improved fidelity and texture realism with flexible data requirements but remain computationally intensive and under evaluation for biomedical applications like virtual staining.

We developed an automated image quality assessment pipeline that integrates multiple metric families to evaluate both image fidelity to evaluate virtual IHC images (\textbf{Figure \ref{fig:graphical_abstract}C}) and staining accuracy (\textbf{Figure \ref{fig:graphical_abstract}D}). Fidelity measures how closely generated images resemble real counterparts, using metrics that quantify latent feature distributions,coverage, diversity, and texture similarity. Metrics in virtual IHC images are benchmarked against real IHC images and pathologist evaluations. We also use simple segmentation metrics such as DICE score, intersection over union (IoU), Hausdorff distance (HD), true positive rate (TPR), and true negative rate (TNR) to measure stain accuracy. As a use case, we selected virtual CDX2 staining due to the expression of CDX2 in glands from the colon.  CDX2 is a nuclear marker of enterocyte differentiation, and prior work demonstrated its utility in generating automated gland outlines \cite{kataria2023automating}. Consequently, CDX2 serves as an ideal target for developing a comprehensive dataset to train and validate virtual staining models.

\begin{figure*}[!thb]
    \centering
    \includegraphics[trim={0.0cm 0.1cm 1.9cm 0.0cm}, clip=true,width=1.0\linewidth]{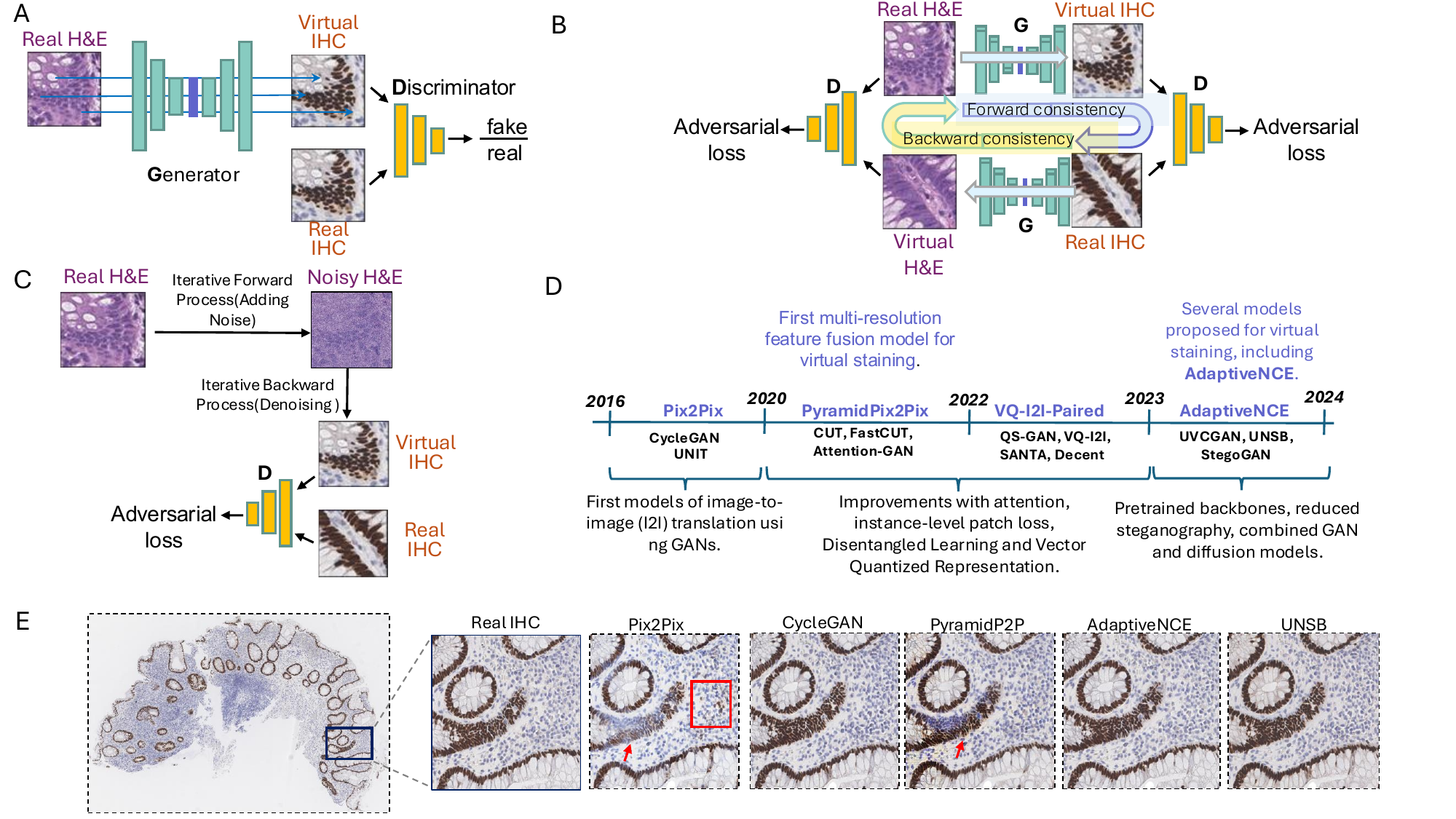}
    \caption{\textbf{Generation of virtual IHC images.} \textbf{A.} Pix2Pix models  predict which pixels in the H\&E tile should be colored. During training, the discriminator decides whether the IHC image is real or a virtual/fake. When the discriminator can no longer distinguish between real and fake IHC, the algorithm completed its training. \textbf{B.} The cycle-GAN architecture uses unpaired image tiles. It includes two discriminator modules, one for real versus virtual IHC images and the other for real versus virtual H\&E images. The consistency loss allows the model to learn from unpaired data. \textbf{C.} The diffusion model uses a GAN architecture to generate virtual IHC images. The Unpaired Neural Schrödinger Bridge (UNSB) model captures continuous, interpretable transitions between H\&E and IHC domains.  It scales to high-resolution biomedical images and supports incorporation of biological priors and regularization. \textbf{D.} Timeline of models for H\&E to IHC image translation. \textbf{E.} Representative examples of generated IHC image tiles. The red arrow points to an area of understaining (false negative pixels) and the red box to an area of overstaining (false positive pixels).}
   
    \label{fig:figure_two} 
\end{figure*}

\paragraph{Generative AI architectures for virtual IHC.} We benchmarked multiple model architectures, including the Pix2Pix family (\textbf{Figure \ref{fig:figure_two}A}), CycleGAN family (\textbf{Figure \ref{fig:figure_two}B}), and its diffusion variant (\textbf{Figure \ref{fig:figure_two}C}), to capture key advances in image translation and generative AI since 2016. All models were trained on the same dataset to compare their performance. Over the past decade, virtual staining methods have progressed rapidly (\textbf{Figure \ref{fig:figure_two}D}). Early image-to-image translation models (2016–2020), based on GAN architectures \cite{isola2017image,zhu2017unpaired,rivenson2019phasestain,rivenson2019virtual}, demonstrated the feasibility of generating synthetic stains but were limited in resolution and staining accuracy. Between 2020 and 2022, the introduction of PyramidPix2Pix \cite{liu2022bci} marked the first model specifically designed for virtual staining of Human Epidermal Growth Factor Receptor 2 (HER2), incorporating multi-resolution feature fusion. Other advances during this period focused on improving pixel-wise reconstruction, integrating attention mechanisms \cite{tang2021attentiongan}, and adopting instance-level contrastive or patch-based losses \cite{park2020contrastive}. From 2022–2023, innovations including query-based attention, disentangled learning, and vector-quantized latent translation \cite{hu2022qs,chen2022eccv} enhanced realism and stability. Most recently (2023–2024), pretrained backbones, hybrid GAN-diffusion frameworks \cite{torbunov2023uvcgan,ho2025f2fldm,kim2023unpaired}, and artifact reduction techniques \cite{wu2024stegogan,kataria2024staindiffuser} have enabled more robust and generalizable virtual staining.

Despite methodological differences, all models adopt an adversarial framework for training: a discriminative network is trained to differentiate real from generated IHC image tiles, while the generative model iteratively improves until the discriminator can no longer reliably distinguish between them. For clinical use, however, success depends on more than adversarial performance. Pathologists require models that produce accurate staining, since misrepresentation of protein expression in IHC images could directly impact diagnostic accuracy and treatment decisions. Representative image tiles from trained image generation models are shown in \textbf{Figure \ref{fig:figure_two}E}. Pix2Pix and Pyramid-Pix2Pix miss positive nuclear staining in glands, while in addition, Pix2Pix falsely stains cells in the lamina propria, i.e., the tissue between glands. Additional WSI predictions from different model families are shown in \textbf{Supplementary Figures} \ref{fig:visual_wsi_1} and \ref{fig:visual_wsi_2}.  

\begin{table*}[!htb]
    \centering
    \begin{tabular}{ccc||cccc}
 \rowcolor{gray}\bf Model  & \bf Architecture & \bf Paired/Unpaired &   \bf FID $\downarrow$ & \bf KID $\downarrow$ & \bf Dist. Precision $\uparrow$ & \bf Dist. Recall $\uparrow$\\
    \hline
\rowcolor{lightgray} Pix2Pix\cite{isola2017image}                            & Pix2Pix & Paired &   11.09    & 0.0064     & 0.8326     & 0.7418\\
\rowcolor{white} PyramidPix2Pix \cite{liu2022bci}   & Pix2Pix & Paired &  25.67       &	0.0248     &	0.6953  &	0.6262\\
\rowcolor{lightgray} AdaptiveNCE \cite{li2023adaptive}  & GAN & Paired & \bf 4.70       &	 \bf 0.0014    	   & \bf 0.9081     &  \bf 0.8594\\
\rowcolor{white} VQ-I2I-Paired  \cite{chen2022eccv} & GAN & Paired & 19.10 &	0.0127	&	0.7193	& 0.5571\\
    \hline 
    \hline
\rowcolor{lightgray} CycleGAN \cite{zhu2017unpaired}  & GAN & Unpaired &  5.04       &  0.0015      &  0.8518 & 0.8382 \\
\rowcolor{white} CUT       \cite{park2020contrastive} & GAN & Unpaired    & 4.64   & 0.0014		   & 0.8549     &  0.8300\\
\rowcolor{lightgray} FastCUT    \cite{park2020contrastive} & GAN & Unpaired   &  8.45      &  0.0054	  	   & 0.8312     &  0.7923\\
\rowcolor{white} Attention GAN  \cite{tang2021attentiongan}  & GAN & Unpaired     & 5.12 & 0.0015	&	0.8618 &	0.8356\\
\rowcolor{lightgray} Decent GAN    \cite{xieunsupervised}    & GAN & Unpaired  & 4.60 &	0.0009	&	0.8647 &	0.8266  \\
\rowcolor{white} QS-GAN \cite{hu2022qs}         & GAN & Unpaired& 4.78	& 0.0016	& 0.8639	& 0.8336\\
\rowcolor{lightgray} UNIT \cite{liu2017unsupervised} & GAN & Unpaired&  6.89 & 0.0028	&0.8484	& 0.7887 \\
\rowcolor{white} SANTA \cite{xie2023unpaired} & GAN & Unpaired&  \bf 4.39 & 	\bf 0.0012	& \bf 0.8756	& \bf 0.8454  \\
\rowcolor{lightgray} VQ-I2I \cite{chen2022eccv} & GAN & Unpaired& 17.77 &	0.0101	&	0.7048	& 0.5087\\
\rowcolor{white} UVCGAN    \cite{torbunov2023uvcgan} & GAN & Unpaired    & 16.89 & 0.0108  & 0.7361 & 0.7125 \\
\rowcolor{lightgray} StegoGAN  \cite{wu2024stegogan}  & GAN & Unpaired   &  10.40 & 0.0072  & 0.7853 & 0.8065 \\
\rowcolor{white} UNSB      \cite{kim2023unpaired}  & Diffusion & Unpaired   &  22.10 & 0.0181 & 0.7006 & 0.6833 \\
    \end{tabular}
    \caption{\textbf{Evaluation of virtual images using standard feature distribution metrics}. The similarity between real and virtual images is assessed using Fréchet Inception Distance (FID) \cite{heusel2017gans}, Kernel Inception Distance (KID) \cite{binkowski2018demystifying}, and feature distribution precision and recall \cite{kynkaanniemi2019improved} evaluated over the full dataset. Distribution precision and recall quantify the diversity and coverage of features in virtual IHC images relative to real images, while FID and KID measure the distance between the two feature distributions. 
    }
    \label{tab:image_realism_metrics}
\end{table*}

\begin{table}[!htb]
    \centering
    \scalebox{1}{
    \begin{tabular}{c||ccc}
  \rowcolor{gray}    \bf Model           & \bf SSIM $\uparrow$ & \bf MSE $\downarrow$& \bf PSNR $\uparrow$ \\        
                 \hline
                 
 \rowcolor{lightgray}   Pix2Pix     &  0.4969  & 235.18 &  19.98 \\
\rowcolor{white} \bf PyramidPix2Pix     &  0.5556  &  224.57 &   20.53 \\
\rowcolor{lightgray} AdaptiveNCE   &  \bf 0.5844 $^{*}$  & \bf 221.97$^{*}$ & \bf 21.00$^{*}$  \\
    \rowcolor{white} VQ-I2I-Paired   & 0.3678 &  256.13 & 17.95 \\
    \hline 
    \hline
  \rowcolor{lightgray}    CycleGAN    &  \bf 0.5502 $^{*}$  & \bf 229.84 $^{*}$ & 19.68 $^{*}$  \\
    \rowcolor{white}  CUT   &   0.5379  &  230.89 & 19.35  \\
    \rowcolor{lightgray} FastCUT   &  0.5280  &  233.52 & 18.66  \\
    \rowcolor{white} Attention GAN     & 0.5458 & 230.31 & 19.60\\
    \rowcolor{lightgray} Decent GAN        & 0.5368 & 233.59 & 19.68\\
    \rowcolor{white} QS-GAN       & 0.5365$^{*}$ & 230.34 & \bf 20.33$^{*}$ \\
    \rowcolor{lightgray} UNIT    & 0.5084 & 254.68 & 18.49 \\
   \rowcolor{white} \bf  SANTA & 0.5419 & 231.47 & 19.90\\
    \rowcolor{lightgray} VQ-I2I  & 0.3186 & 264.26 & 16.40 \\
    \rowcolor{white} UVCGAN    &   0.5455 &  233.29& 19.62 \\
    \rowcolor{lightgray} StegoGAN   &  0.5465  & 234.90 & 19.31 \\
     \rowcolor{white} UNSB    &   0.5250 & 240.10 & 19.29 \\

    \end{tabular}}
    \caption{\textbf{Evaluation of virtual IHC images using texture metrics.}  PSNR (Signal to Noise Ratio), SSIM (Structural Similarity Index) and MSE (Mean Square Error) represent the average values of tiles. $^{*}$ indicates a statistically significant difference between the best and the second best performing model (p $<$ 0.005):  AdaptiveNCE\cite{li2023adaptive} vs. PyramidPix2Pix  for all metrics, CycleGAN vs. StegoGAN for SSIM, CycleGAN vs. AttentionGAN for MSE, and QS-GAN vs. SANTA for MSE. }
    \label{tab:texture_metrics}
\end{table}

 \begin{figure*}[!htb]
    \centering
    \includegraphics[width=1.0\linewidth]{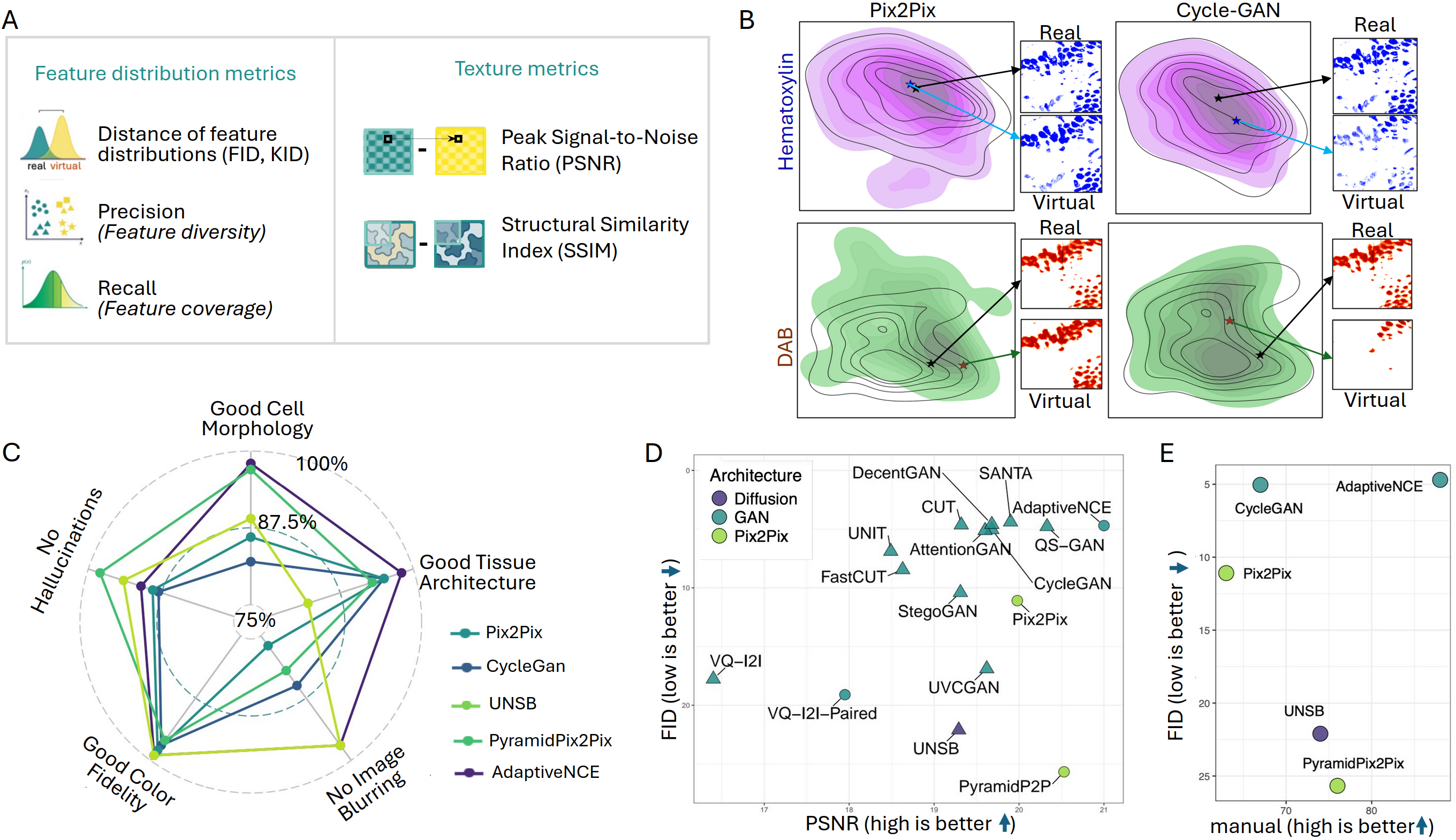}
    \caption{\textbf{Conventional metrics for evaluation of image quality.} \textbf{A.} Metrics categories: feature distribution metrics evaluate features that are generated by encoders of real and virtual images. Texture metrics evaluate pixel-wise differences between paired real and virtual images. \textbf{B.} Hematoxylin and DAB feature coverage in real and virtual images. The hematoxylin and DAB channels of tiles are unmixed and passed through the same encoder. The area of solid color depicts the feature densities of virtual images while the dashed lines show the feature densities of real images. The image tiles on the side are added for  qualitative comparisons of real and virtual images. \textbf{C.} Manual evaluation of image tiles generated by five models. The percentage of image tiles with good cell morphology, good tissue architecture, no blurring, good color fidelity and no hallucinations is shown. \textbf{D.} Comparison of FID scores and average PSNR scores. Models using unpaired input data are shown by triangles and models using paired inputs by circles. \textbf{E.} Comparison of FID scores and manual quality metrics.  }
    \label{fig:figure_three}
\end{figure*}

\paragraph{Image fidelity evaluation of virtual staining models.}  Feature distribution metrics (\textbf{Figure \ref{fig:figure_three}A}), are computed from distributions of high-dimensional features that are generated by image encoders( foundation or pretrained models)  \cite{szegedy2015going,dubey2023structural,kataria2024staindiffuser,liu2022bci,torbunov2023uvcgan}. The Frechet Inception Distance (FID) score \cite{heusel2017gans} models the high dimensional feature as Gaussian distributions. By estimating the mean and covariance for real ($\mu_{real},\Sigma_{real}$) and virtual ($\mu_{virtual},\Sigma_{virtual}$) IHC images, the FID score reports the distance between the distributions. The Kernel Inception Distance (KID) uses a polynomial kernel instead of a Gaussian. In addition, precision and recall quantify how well features from virtual images fall within the distribution of features from real images. The precision and recall in the latent feature space corresponds to feature diversity and coverage, respectively. In contrast to distribution metrics, texture metrics (PSNR, MSE, SSIM) measure average pixel deviations, thereby evaluating differences in luminance, contrast, and structure (\textbf{Figure \ref{fig:figure_three}A})  \cite{dubey2023structural,dubey2024vims,kataria2024staindiffuser,li2023adaptive,liu2022bci,bian2024hemit}.

FID and KID scores are lower for models trained on unpaired compared to paired data, consistent with a better performance of unpaired models (\textbf{Table \ref{tab:image_realism_metrics}}). A notable exception is AdaptiveNCE \cite{li2023adaptive}, which after training on paired image tiles matches the top unpaired model, SANTA \cite{xie2023unpaired}, in FID and KID scores. AdaptiveNCE also achieves superior feature coverage and diversity, as indicated by its high precision and recall in latent feature space. Across models, we observe strong correlations between FID and KID scores ($r=0.97$)\footnote{$r$ is the Pearson correlation coefficient.}, FID and distribution precision ($r=-0.96$), and FID and distribution recall ($r=-0.86$), suggesting that these metrics capture overlapping image quality attributes. Taken together, these results highlight that while unpaired training often leads to better fidelity of feature distributions, certain models, such as AdaptiveNCE when trained on paired data can match or exceed unpaired methods. In addition to the standard FID scores obtained from the ImageNet-pretrained Inception encoder, we computed distribution metrics using two pathology-trained encoders, UNI and UNI-2, to assess the impact of domain-specific high dimension feature representation. While the FID and KID values are higher with the pathology-trained compared to ImageNet-pretrained encoders, AdaptiveNCE and SANTA reveal the lowest scores with both types of encoders. The scatter plots in \textbf{Supplementary Figure} \ref{fig:fid_fid_uni} and corresponding values in \textbf{Supplementary Table} \ref{tab:pathology_domain_distribution_metrics} reveal a strong correlation between FID-Inception and FID-UNI ($r = 0.9258$, \ref{fig:fid_fid_uni}A) as well as FID-Inception and FID-UNI-2 ($r = 0.8042$, \ref{fig:fid_fid_uni}B). These findings suggest that histopathology-specific encoders capture distributional information comparable to ImageNet-trained encoders when evaluating virtual IHC images \cite{piatrikova2025modelling}.  Therefore, we adopt Inception-Net FID as the baseline distribution metric throughout this paper.

To compare the feature coverage of real and virtual IHC images, we used a 2D UMAP (\textbf{Figure \ref{fig:figure_three}}B). To separately evaluate hematoxylin (blue) and DAB-stain (brown) features we unmixed blue and brown pixels before UMAP projections. Pix2Pix-generated images exhibit broader feature coverage in both hematoxylin and DAB channels compared to real IHC, whereas CycleGAN features show more restricted coverage. CycleGAN features in hematoxylin and DAB channels show greater overlap with real IHC features in parallel with the higher distribution recall values in \textbf{Table} \ref{tab:image_realism_metrics}. Notably, in the DAB channel, the UMAP distance between real and virtual Pix2Pix and CycleGAN images is similar, yet the virtual Cycle-GAN image misses brown nuclei. To further determine the interpretability of the 2D UMAP distance, we compared the UMAP distance to the distance between the same virtual and real image tiles in the high-dimensional vector space (\textbf{ Supplementary Figure \ref{fig:UMAP-L2}). } A modest correlation (\textit{r} = 0.6) indicates that 2D UMAP distances should be interpeted with caution as a measure of tile-wise feature similarities.

In contrast to distribution-based metrics, values of texture metrics indicate that paired models consistently outperform unpaired models (\textbf{Table \ref{tab:texture_metrics}}). Notably, the paired VQ-I2I \cite{chen2022eccv} surpasses its unpaired counterpart, further emphasizing the advantage of paired approaches for texture preservation. Compared to unpaired models, paired models achieve better texture metrics, particularly PSNR and MSE. Violin plots of tile-wise texture values (\textbf{Supplementary Figure} \ref{fig:texture_voilin}) show a normal distribution of each metric.  Additionally, higher PSNR values are strongly associated with lower MSE ($r=-0.908$) and higher SSIM ($r=0.799$).

Paired models (Pix2Pix, PyramidPix2Pix, AdaptiveNCE) show similar PSNR despite differing FID scores, whereas unpaired models (CUT, SANTA, QS-GAN) have comparable FID but vary in PSNR. Comparing distribution and texture metrics across models, we observed a weak inverse correlation between texture metrics and FID ($r_{psnr,fid}=-0.24$,$r_{ssim,fid}=-0.365$,$r_{mse,fid}=-0.28$ \textbf{Figure \ref{fig:figure_three}D}). These results indicate that distribution- and texture-based metrics capture different attributes of virtual IHC image fidelity.

\textbf{Correlation between image fidelity and manual image quality metrics.} Despite their wide use, distribution- and texture-based metrics have not been systematically benchmarked on virtual IHC images against pathologist evaluations. To determine the relationship of image fidelity with manual evaluations (\textbf{Figure \ref{fig:figure_three}C})  a pathologist, blinded to the models, assigned scores of (i) tissue architecture, (ii) cell morphology, (iii) image blurriness, (iv) color fidelity, and (v) presence of hallucinations to virtual tiles (\textbf{Supplementary Table} \ref{tab:manual_annotation_descritpion}). Only the top performing models were used for manual evaluation. AdaptiveNCE  achieved the highest score overall with the highest percentage of image tiles with perfect cell morphology and tissue architecture, and with minimal image blurriness. PyramidPix2Pix revealed the fewest hallucinations. All models scored high in color fidelity. A rank list of manual image quality assessment revealed AdaptiveNCE (88\% of tiles with good scores) > PyramidPix2Pix (76\%) > UNSB (74\%) > cycleGAN (67\%) and > Pix2Pix (63\%) (\textbf{Figure \ref{fig:figure_three}C}).

\begin{table*}[!htb]
    \centering
    \scalebox{1.0}{
    \begin{tabular}{c||ccccc||ccccc}
    \rowcolor{gray} &  \multicolumn{5}{c||}{\bf DAB mask} & \multicolumn{5}{c}{\bf Segmentation Model} \\
    \hline
    Model     & DICE $\uparrow$ & IOU $\uparrow$ & HD  $\downarrow$ & TPR$\uparrow$ & TNR $\uparrow$ & DICE $\uparrow$ & IOU $\uparrow$ & HD  $\downarrow$ & TPR$\uparrow$ & TNR $\uparrow$\\        
                 \hline
   \rowcolor{lightgray}  \bf Pix2Pix  &  0.75 &  0.65 &  20.72 &  0.74 &  \bf 0.98  &  0.82 & 0.71 & 14.98 &  0.80 &  \bf 0.98\\
    \rowcolor{white}   PyramidPix2Pix  & \bf 0.78$^{*}$ & \bf 0.68$^{*}$ & \bf 20.24 & \bf 0.80$^{*}$ &  0.98$^{*}$  & \bf 0.82$^{*}$ & \bf 0.72$^{*}$ &  \bf 13.15$^{*}$ & \bf 0.84$^{*}$ &  0.97$^{*}$\\
    \rowcolor{lightgray}   AdaptiveNCE    &  \bf 0.78$^{*}$& \bf 0.68$^{*}$& \bf 19.87$^{*}$& \bf 0.80$^{*}$&  0.98$^{*}$& \bf 0.82$^{*}$&  \bf 0.72$^{*}$& \bf 12.89$^{*}$& \bf 0.84$^{*}$ &   0.97$^{*}$ \\
    \rowcolor{white}   VQ-I2I-Paired  &  0.68 & 0.55 & 44.83 & 0.70 & 0.95 &  0.75 & 0.61 & 24.56 & 0.78 & 0.95\\
        
    \hline
    \hline

     \rowcolor{lightgray} CycleGAN  & 0.70 & 0.58 & 29.46 & 0.71 & 0.97 & 0.76 &  0.63 & 19.97 & 0.76 &  0.97 \\

\rowcolor{white}     CUT  &    0.68 & 0.57 & 32.24 & 0.69 & 0.97 & 0.75 &  0.62 & 21.74 & 0.75 &  0.97 \\
    \rowcolor{lightgray} FastCUT  &  0.57 & 0.46 & 45.95 & 0.53 & \bf 0.98 & 0.65 &  0.52 & 38.64 & 0.60&  \bf 0.98 \\
    \rowcolor{white} Attention GAN   & 0.69 & 0.58 & 30.83 & 0.71 & 0.97 &  0.77 & 0.64 & 20.11 & 0.76 & 0.97\\
    \rowcolor{lightgray} \bf Decent GAN      & 0.72 & 0.61& 28.42 & 0.73& 0.97&  0.78& 0.67 & 18.54& 0.79& 0.97 \\
    \rowcolor{white} QS-GAN       &  0.72 & 0.61 & 28.42 & 0.73 & 0.97 &  0.78 & 0.67 & 18.54 & 0.79 & 0.97 \\
    \rowcolor{lightgray} UNIT  &  0.51 & 0.42 & 42.05 & 0.52 & 0.97 &  0.61 & 0.51 & 18.91 & 0.77 & 0.97 \\
    \rowcolor{white} SANTA  & \bf  0.73 & \bf 0.62$^{*}$ & \bf 25.45 & 0.73$^{*}$ & 0.97$^{*}$ & \bf  0.80$^{*}$ & \bf 0.68$^{*}$ & \bf 17.27 &  0.79$^{*}$ &  0.97$^{*}$ \\
    \rowcolor{lightgray} VQ-I2I &  0.54 & 0.42 & 79.79 & 0.58 & 0.92 &  0.62 & 0.49 & 44.98 & 0.66 & 0.93 \\
    \rowcolor{white} UVCGAN     &  0.65& 0.52& 58.70& \bf 0.78& 0.92& 0.68&  0.53 & 35.00 & \bf 0.82 &  0.91 \\
    \rowcolor{lightgray} StegoGAN   &  0.69 & 0.58 & 38.24 & 0.72 & 0.97 & 0.74 &  0.61 & 22.57 & 0.77 &  0.96 \\
     \rowcolor{white} UNSB   &   0.71 & 0.59 & 34.28& 0.73 & 0.97& 0.76 &  0.63 & 20.23 & 0.77 &  0.97 \\
    \end{tabular}}
    \caption{\textbf{Stain accuracy evaluation.} Stain accuracy metrics comparing pixel masks from real and virtual IHC images. Metrics include DICE score, Intersection over Union (IoU), Hausdorff distance (HD), true positive rate (TPR), and true negative rate (TNR). $^{*}$ indicates a statistically significant difference ($p < 0.005$, t-test). \textbf{Pix2Pix} and \textbf{DecentGAN} (second-best models) are compared against the top-performing paired and unpaired models.}
    \label{tab:segmentation_metrics}
\end{table*}

Next, we compare FID scores with manual image quality assessments (\textbf{Figure \ref{fig:figure_three}E}). No correlation was observed between FID and manual fidelity scores ($r=-0.034$); notably, Pix2Pix and CycleGAN, despite low manual scores, achieved better FID than UNSB and PyramidPix2Pix, which were rated higher manually.

Comparing the texture metrics with manual assessments revealed good correlations ($r_{psnr,manual}=0.675$, $r_{ssim,manual}=0.8345$, $r_{mse,manual}=-0.623$) for the model level. However, when analyzed at tile-level the correlation is lower ($r_{psnr,manual}=0.22$, $r_{ssim,manual}=0.23$, $r_{mse,manual}=-0.09$). To further evaluate  metrics at tile-level, we divided image tiles into two groups with good or poor image quality based on manual scoring (\textbf{Figure \ref{fig:figure_three}C}). We compared PSNR, SSIM and MSE values between the two groups using a t-test. UNSB was the only model that showed statistically significant differences ($p<0.05$) across all texture image quality metrics, whereas CycleGAN showed a significant difference only in PSNR; all other comparisons between the two groups were insignificant (\textbf{Supplementary Table \ref{tab:texture_metrics_p_values}}). 
These results indicate that, although texture metrics correlate well with manual assessments at the model level, they are unreliable indicators of tile-level manual assessment. Overall, while texture and distribution metrics capture certain aspects of image fidelity, they do not agree with manual image quality at tile level. This limits the utility of texture metrics for establishing the clinical relevance of virtual IHC stains.

\paragraph{Evaluation of staining accuracy in virtual IHC images using segmentation metrics.} 
Color deconvolution of hematoxylin and DAB \cite{kataria2023automating,ruifrok2001quantification} enables generation of IHC positive pixel masks without the input from a pathologist. This approach has been successfully used in automated annotation frameworks \cite{kataria2023automating,brazdil2022automated,arvaniti2018automated}, and we apply it here to generate stain-positive masks in real and virtual IHC images (\textbf{Figure \ref{fig:figure_four}A}; sample masks in \textbf{Supplementary Figure} \ref{fig:dab_model_segmentation}). We also compare fixed-value and model-based thresholding for mask generation  \cite{kataria2023automating,brazdil2022automated,fischer2008hematoxylin}, \cite{kataria2023automating} (see Methods and Supplementary Methods sections \ref{suppsec:supp:methods}). A comparison of IHC positive pixel masks generated by these two methods is shown in \textbf{Supplementary Figure} \ref{fig:dab_model_segmentation}. While the model outputs are highly correlated ( $r=0.979$, at patch level), the deep learning model generates more stable and consistent binary masks and is 5\% more accurate than threshold based segmentation (\textbf{Table \ref{tab:segmentation_metrics}}). In addition, the violin plots in Supplementary Figure \ref{fig:model_segmentation_voilin} reveal a larger number of tiles without DAB masks when using fixed-value thresholding, highlighting the sensitivity of this approach.
\begin{figure*}[!htb]
    \centering
    \includegraphics[width=1.0\linewidth]{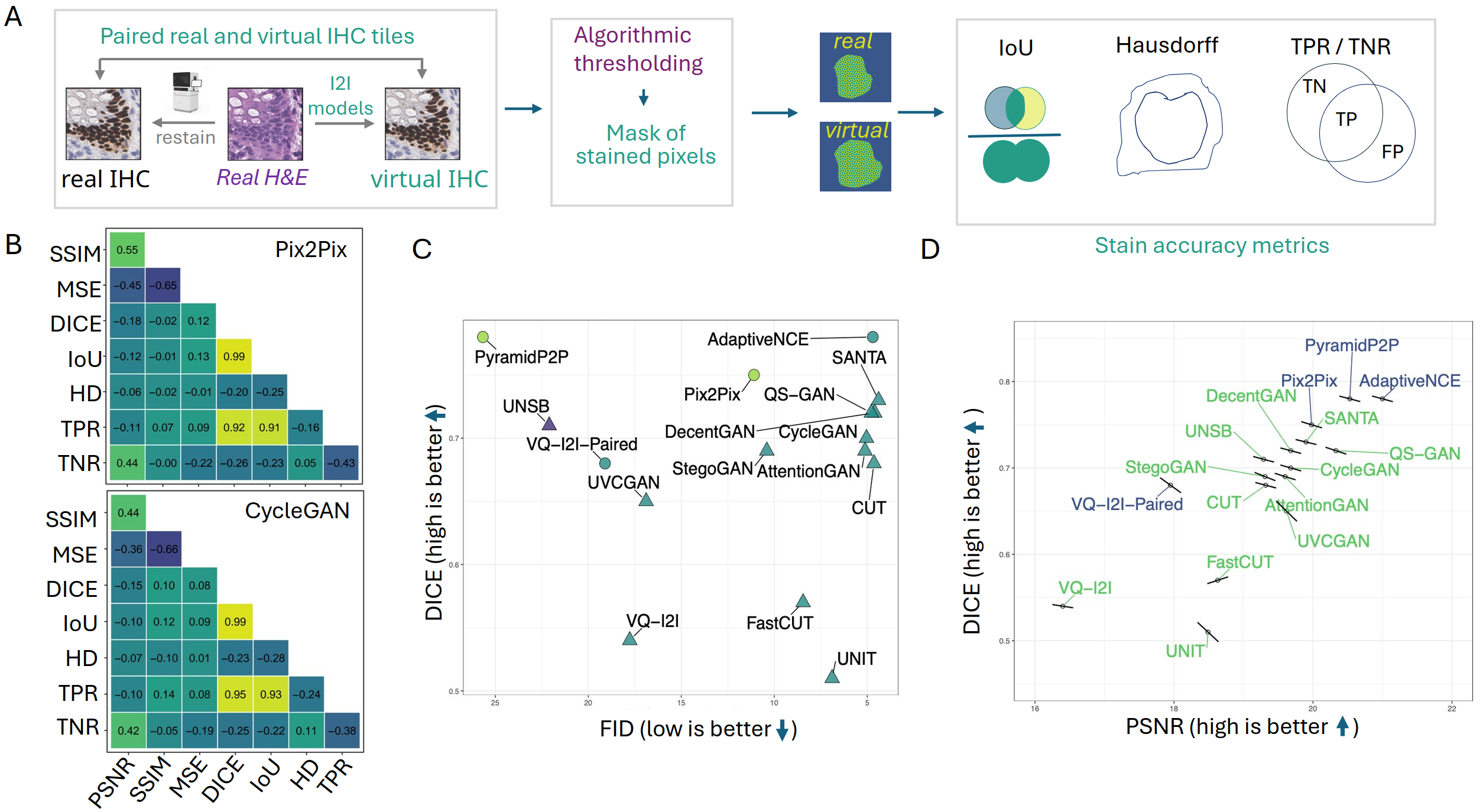}
    \caption{\textbf{Metrics for evaluation of staining accuracy.} \textbf{A.} Workflow to determine staining accuracy. After digitization the H\&E-stained slides, the tissue is restained with the CDX2 antibody and DAB as the chromogen. Alternatively, the digital H\&E tiles are used to generate virtual CDX2-IHC tiles. Real and virtual IHC tiles are registered at pixel level accuracy. The brown color IHC stain in real and virtual IHC image tiles is converted to a binary DAB pixel mask using a trained model. After registration the DAB mask in the virtual tile is compared to the DAB mask in the real tile using the stain accuracy metrics of IoU, DICE and Hausdorff distance (HD). True positive (TPR) and true negative rates (TNR) are calculated in addition to false positive (FP) and false negative pixel rates. \textbf{B.} Tile-wise correlations between texture metrics (PNSR, SSIM, MSE) and stain accuracy metrics (IoU, DICE, HD, TPR, TNR) in Pix2Pix and CycleGAN models. \textbf{C.} Comparison of average DICE with FID scores. Models using unpaired input data are shown by triangles and models using paired input data by circles. \textbf{D.} Comparison of average DICE and average PSNR scores. Circles indicate the average scores across all the tiles in the dataset.  Black lines show the relationship of patch-wise DICE and PSNR scores within each model. Note the negative regression slopes of tile-wise DICE and PSNR scores within each model in contrast to the strong positive correlation of average DICE and average PSNR between models. PSNR - Peak Signal-to-Noise Ratio, SSIM – Structural Similarity Index, MSE- Mean Square Error, DICE – DICE Similarity Coefficient, IoU – Intersection over Union, HD – Hausdorff Difference.}
    \label{fig:figure_four}
\end{figure*}

Image segmentation metrics (\textbf{Table} \ref{tab:segmentation_metrics}) show that paired models achieve higher stain accuracy compared to unpaired models.  PyramidPix2Pix and AdaptiveNCE perform nearly identically, differing only slightly in Hausdorff distance. Amongst the unpaired models SANTA performs best. 
Both paired and unpaired models consistently achieve high true negative rates (TNR), but struggle with accurate IHC-positive staining (TPR).
Pix2Pix and CycleGAN reveal strong positive pairwise correlations between DICE, IoU, and TPR ($r > 0.90$), weak negative correlations of these metrics and Hausdorff distance ($r \approx -0.25$ to $-0.3$). (\textbf{Figure \ref{fig:figure_four}B}; correlation maps for all other models in \ref{fig:correlation_maps}).

\paragraph{Comparison of image fidelity and image accuracy metrics. } To assess whether image fidelity metrics correlate with stain accuracy, we compared FID and mean values of texture metrics against mean DICE scores (\textbf{Figure \ref{fig:figure_four}}). Across all models, the Pearson correlation between FID and DICE is close to zero ($r = 0.002$), confirming quantitatively that distribution metrics do not reliably measure staining accuracy (\textbf{Figure \ref{fig:figure_four}C}).

We calculated the correlation between texture metrics values and DICE scores at model and patch levels.  To obtain correlation coefficients at the model level, we used the average measurements of DICE and PSNR, SSIM and MSE across all tiles from each of the 16 models.  Pairwise correlations coefficients between DICE and each of the texture metrics are in the moderate to high range ($r_{psnr,DICE}=0.80$,  $r_{ssim,DICE}=0.481$,  $r_{mse,DICE}=-0.6629$) (\textbf{Figure} \ref{fig:figure_four}C). However, when inspecting the slopes for each of the models that are generated by tile-wise measurements, we observed small or opposite relationships between DICE and the texture metrics (\textbf{Figure} \ref{fig:figure_four}D). At patch level, correlations of  DICE, IoU \& HD with PSNR, SSIM, MSE  were near zero or negative (\textbf{Figure \ref{fig:figure_four}B}, Supplementary Figure \textbf{\ref{fig:correlation_maps}} ).  These results show a disagreement between tile-level and model-level evaluations when considering texture metrics.

\begin{figure*}[!htb]
    \centering
     \includegraphics[width=1.0\linewidth]{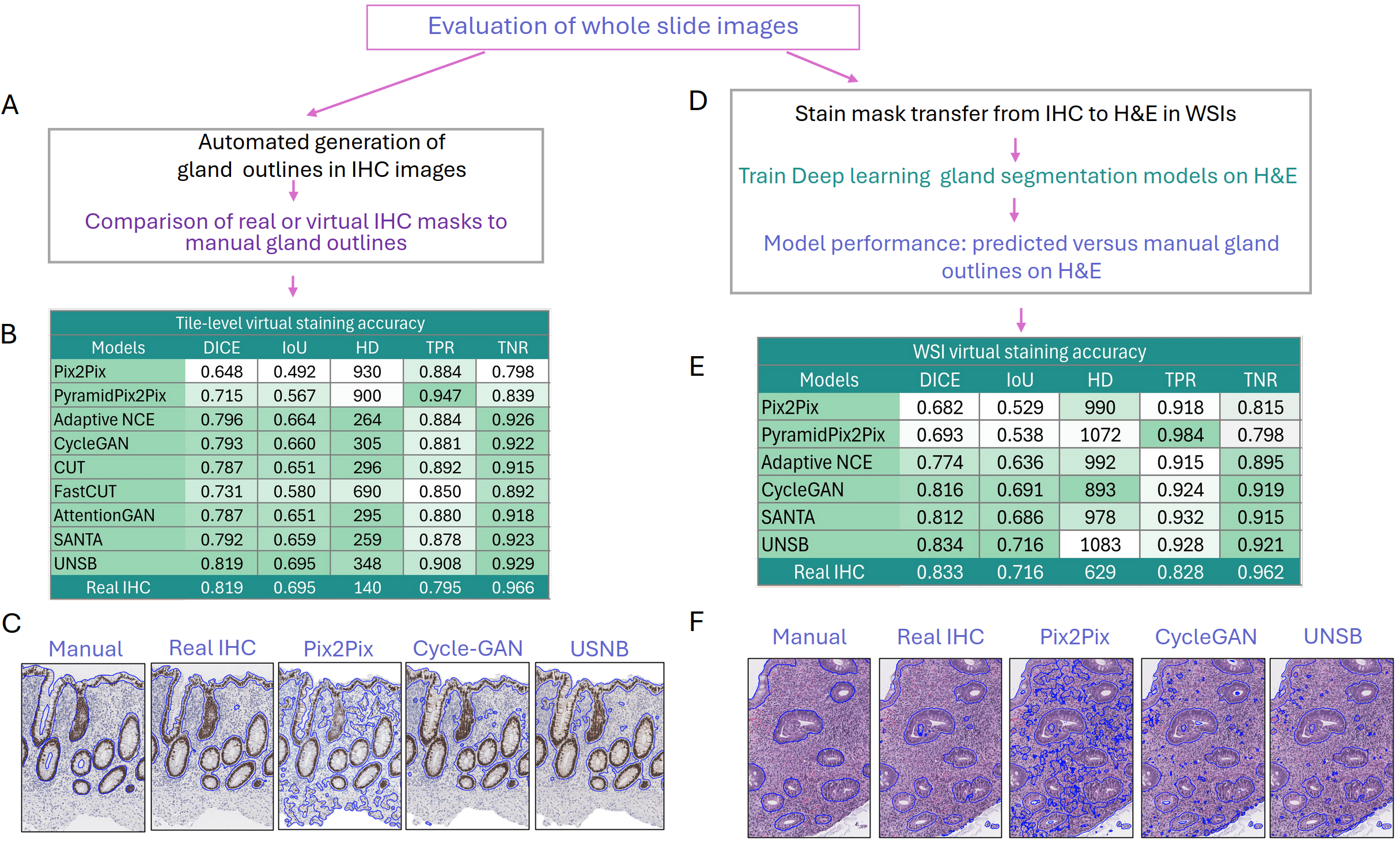}
\caption{\textbf{Stain accuracy evaluation in WSIs}. \textbf{A - C}. \textit{Accuracy of gland segmentation in virtual IHC WSI.} Comparison of automated gland masks in real and virtual IHC WSI to manual gland annotations. A trained algorithm is used to generate gland outlines from the IHC pixel masks.  \textbf{B.} Gland outlines based on virtual IHC masks from the models listed in the first column are compared to manual IHC outlines using DICE, IoU, HD, TPR and TNR metrics. For comparison, the results of the real IHC gland outlines are shown in the bottom row. \textbf{C.} Qualitative evaluation of gland segmentation. In the left tile, the glands are outlines by a pathologist. Note the difference in false positive annotations in the lamina propria outside the glands in the real/virtual IHC images. \\ \textbf{D - F.} \textit{Performance of model trained on H\&E gland segmentations}. The DAB pixel masks in virtual or real IHC images are transferred to the corresponding H\&E image. Annotations in the H\&E image are used to train gland segmentation models. \textbf{E}. The performance of the gland segmentation models in a held-out test set is compared to a model trained on transferred real IHC gland outlines. Metrics as in B. \textbf{F.} Qualitative segmentation results of gland outlines generated by models trained directly on manual H\&E gland outlines, gland outlines transferred from real IHC images and gland outlines transferred from virtual IHC images. The models used to generate the virtual IHC images are listed above the image. }
    \label{fig:wsi_evaluation}
\end{figure*}

\paragraph{Assessment of stain accuracy in whole slide virtual staining predictions.} In the current virtual staining literature \cite{li2023adaptive,liu2022bci,kataria2024staindiffuser,kataria2025implicitstainer}, image quality evaluations are performed using patch-based frameworks rather than whole-slide images (WSIs, which often exceed 10,000 × 10,000 pixels in size). We therefore questioned whether stain accuracy metrics defined over patches/tiles are suitable for assessing the quality of WSIs. To this point, we directly compared real and virtual CDX2-based gland outlines to manual outlines as ground truth in WSIs \cite{kataria2023automating,brazdil2022automated} (\textbf{Figure} \ref{fig:wsi_evaluation}A-C). We also transferred the real and virtual gland outlines to H\&E images and trained a separate gland segmentation model for each virtual staining model  \cite{kataria2023automating} (\textbf{Figure} \ref{fig:wsi_evaluation}D-E). We observed a marked decrease in model performance by block artifacts in reconstructed WSIs (\textbf{Supplementary Figure} \ref{fig:block_artifacts}). To mitigate this problem, we used a sliding window with a 192-pixel overlap. However, residual artifacts remained, with stain-positive regions showing signs over-smoothing. Smoothing generates a loss of fine-grained structural detail because the model averages across outputs from adjacent tiles instead of committing to sharp transitions. The problem is particularly pronounced in Pix2Pix \cite{isola2017image} and PyramidPix2Pix \cite{liu2022bci} and leads to a decline in DICE scores of gland outlines by 20\% (\textbf{Figure} \ref{fig:wsi_evaluation}).

Notably, the gland outlines in virtual WSIs generated by UNSB, AdaptiveNCE, and CycleGAN models closely approximate those observed in real IHC images. However, even in these models, a decline in accuracy between tile and WSI gland outlines is observed using the HD metric  (Table \ref{tab:segmentation_metrics} vs. Figure \ref{fig:wsi_evaluation}B). Altogether, masks in  WSI are less reliable (\textbf{Figure \ref{fig:wsi_evaluation}C}; \textbf{Supplementary Figure~\ref{fig:visual_wsi_1}}), particularly in regions with pronounced block artifacts. Visual inspection (\textbf{Figure \ref{fig:wsi_evaluation}C}) further shows that paired models often generate falsely stained cells in the lamina propria between glands, reducing the true negative rate (TNR) (\textbf{Supplementary Figures \ref{fig:visual_wsi_1}} and \textbf{\ref{fig:visual_wsi_2}}). Since most virtual staining methods do not account for WSI-scale processing, our framework is the first to systematically highlight these additional challenges using quantitative metrics. Paired models, except AdaptiveNCE, are particularly prone to tiling artifacts, whereas unpaired models are more robust.

\textbf{WSI stain accuracy drives downstream application success}. Using the automated annotation pipeline described in Kataria et al.\ \cite{kataria2023automating}, we transferred gland annotations from IHC images to their corresponding H\&E WSIs and used these H\&E annotations to train separate gland segmentation models on transferred gland outlines from each of the virtual IHC models (\textbf{Figure \ref{fig:wsi_evaluation}D}). The trained models were evaluated using a held-out set of WSIs with manual gland annotations. DICE, IoU, HD, TPR, and TNR evaluation results are shown in (\textbf{Figure \ref{fig:wsi_evaluation}E}). As anticipated, models trained on higher-quality virtual IHC WSIs consistently achieve better  segmentation performance as shown in \textbf{Figure \ref{fig:wsi_evaluation}F}. UNSB achieved the same DICE and IoU values as real IHC, however, a poorer HD value.  In contrast, models trained on paired data exhibited boundary inaccuracies and regions with false-positive cell coloring. Overall, the data demonstrate that the quality of virtual staining directly influences the effectiveness of downstream tasks, underscoring the need for a rigorous and quantitative framework for evaluating virtual IHC performance on WSIs.

\section*{Discussion}

This study introduces an automated and scalable framework for evaluating the quality of virtual immunohistochemistry (IHC) images. Benchmarking sixteen paired and unpaired image-translation models, we show that conventional fidelity metrics, such as FID, PNSR and SSIM correlate poorly with pathologist assessment and staining accuracy at the patch level. Segmentation-based measures from automatically generated stain masks offer more reliable  and interpretable metrics of whether models correctly label IHC-positive cells. These results highlight the limitations of commonly used metrics \cite{li2023adaptive,liu2022bci,kataria2024staindiffuser} and underscore the need for assessing both visual fidelity and staining accuracy to determine the quality of generated IHC stains. 

\paragraph{Image fidelity versus stain accuracy.} Our findings underscore a key conceptual distinction: \textit{good image fidelity does not equal stain accuracy}. Generative models are typically optimized using perceptual loss functions that favor texture fidelity, yet these do not necessarily correspond to correct IHC positive pixel labeling based on cell lineage, differentiation, or activation states. Similar observations have been reported in virtual fluorescence labeling and cross-modality image synthesis, where outputs may appear convincing but misrepresent the underlying biological signals \cite{thummerer2023synthrad2023}. In the context of virtual IHC, such discrepancies manifest as false-positive or false-negative cells that conventional metrics fail to detect \cite{dubey2023structural}. While FID and KID remain widely used for assessing generative models \cite{liu2022bci,li2023adaptive,kataria2024staindiffuser}, they capture differences in abstract feature distributions rather than true staining correctness. Texture-based metrics like PSNR and SSIM quantify low-level image similarity but cannot reliably reflect molecular or cellular fidelity. Domain-specific adaptations, such as perceptual losses trained on histology encoders or cell-level fidelity indices, provide no improvements as they still rely on proxy measures rather than a direct comparison at the pixel level to determine correct cell staining  \cite{chen2024uni,vorontsov2023virchow,kataria2023pretrain,zimmermann2024virchow2} . Importantly, our analysis shows that although texture metrics moderately correlate with pathologist-informed assessments, they are insufficient as stand-alone indicators of stain accuracy at tile level. However, when comparing averaged texture metric values and stain accuracy across models, a  moderate to high correlation was observed. The discrepancy in tile-wise and model-wise relationship reveals a challenge in comparing different types of metrics. Furthermore, the primary differences in image fidelity and stain accuracy metrics arise between models trained on paired versus unpaired input data, rather than from variations in model complexity or architectural design (e.g., GAN-based vs. diffusion-based).

The importance of image fidelity and staining accuracy depends on the use of virtual IHC images. Current I2I models faithfully recapitulate tissue architecture and nuclear morphology—features dominated by hematoxylin contrast, as shown by our pathologist evaluation—but they remain less reliable in correctly labeling stain positive cells.  The stain accuracy may further decline for markers expressed in cell types without distinctive morphology on H\&E, such as specific T-cell subsets or macrophage populations. Consequently, different clinical and research applications will impose different thresholds for acceptable image quality: virtual IHC intended for direct human interpretation demands both high image fidelity and stain accuracy, whereas tasks such as downstream treatment-response prediction that rely on aggregate statistics (e.g., percent positive cells\cite{dubey2023structural}) can tolerate modest inaccuracies at the single-cell level. Overall, image quality assessment must therefore be contextualized by the intended downstream task, rather than treated as a task-agnostic property of the generative model. 

\paragraph{\textbf{Pathologist-free quantitative assessment of staining accuracy.}} To automate the evaluation of stain accuracy, we restain H\&E-stained tissues with CDX2, a cell lineage specific antibody. This approach eliminates the need for manual pathologist annotations and is generalizable across antibodies and tissue types. Overall, the proposed image quality evaluation framework enables reliable assessment at both tile and WSI levels thus allowing scalable evaluations that are difficult to obtain through manual generation of ground truth. By combining accuracy and fidelity metrics, our framework supports robust, high-throughput benchmarking of generative models. Accuracy-based evaluations can advance virtual staining by promoting reliable and trustworthy adoption of virtual staining by pathologists in clinical workflows.

\paragraph{\textbf{WSI evaluations.}} Extending our evaluation to WSIs highlights additional challenges. Conventional virtual staining assessments are typically performed on image tiles.  Methods to generate WSIs from individual tiles compromise staining accuracy at tile boundaries  \cite{li2023adaptive,liu2022bci,zhou2025protomtg,klockner2025gans}. While some preliminary studies have attempted to mitigate these issues using tile-consistency losses \cite{lahiani2020seamless,liu2025generating}, these approaches offer only partial solutions. The challenge of maintaining spatial and structural consistency across large-scale images remains unresolved. Addressing this limitation will require novel image translation architectures and generative methodologies capable of modeling long-range dependencies, minimizing tiling artifacts, and ensuring accurate staining across entire tissue regions.

\paragraph{Limitations.} Our study has several limitations that suggest directions for future research. First, segmentation-based metrics depend on pixel-level paired annotations, and their applicability to unpaired datasets remains uncertain. For unpaired datasets, directly using staining accuracy is not feasible due to the absence of ground truth; therefore, an appropriate consolidation between accuracy and fidelity metrics is necessary for reliable evaluation. Additionally, building paired datasets is resource-intensive, highlighting the need for protocols or methods leveraging unpaired data. Second, our analysis focuses on brightfield images; future work should assess other virtual staining tasks, including unstained-to-stained and autofluorescence-based image generation, which pose distinct challenges.

Finally, we did not perform a human evaluation in which virtual IHC replaces real IHC in clinical workflows, the gold standard for assessing clinical validity. Our goal was instead to establish an automated framework for evaluating staining accuracy in scenarios where manual assessment is unavailable, unreliable, or impractical. For example, in CD3 staining, manual annotation is unreliable because T-cells cannot be distinguished from other lymphocytes in H\&E-stained tissue. In such cases, our approach provides the only feasible means of assessing virtual staining performance.

\section*{Conclusion and Future Work}

Hematoxylin and Eosin staining remains a cornerstone of pathology; however, in certain cases, it fails to provide the molecular or structural information necessary for accurate diagnosis. IHC  provides critical insights into cell lineage, differentiation, and activation states, yet it is time-consuming, costly, and requires specialized infrastructure. Virtual staining with deep learning offers a scalable and cost-effective alternative. Here, we present a comprehensive framework demonstrating that commonly used distribution- and texture-based metrics—such as FID, KID, PSNR, and SSIM—do not accurately capture correctness of cell staining. In contrast, segmentation–based metrics derived from pixel-level paired annotations provide interpretable, scalable measures of true cell staining. Whole-slide image evaluations further reveal challenges overlooked in tile-level analysis, emphasizing the need for metrics that account for both stain accuracy and image fidelity at WSI level. We envision this framework serving as a practical guideline for researchers developing and benchmarking virtual staining models, enabling the systematic comparison of new architectures and methodological improvements. Future work will aim to extend this framework to unpaired datasets and reduce artifacts when applying novel virtual staining approaches to WSIs. By addressing these challenges, we hope to accelerate the translation of virtual staining technologies into reliable, clinically actionable tools that can support pathologists in improving diagnostic accuracy and patient care.

\section*{Methods}

\paragraph{Image Translation Methodology Details.} 
In our benchmark, we include four paired translation models: Pix2Pix \cite{isola2017image}, PyramidPix2Pix \cite{liu2022bci}, AdaptiveNCE \cite{li2023adaptive}, and VQ-I2I \cite{chen2022eccv} and twelve unparied translation models, which adopt either GAN-based frameworks—CycleGAN \cite{zhu2017unpaired}, CUT \cite{park2020contrastive}, FastCUT \cite{park2020contrastive}, AttentionGAN \cite{tang2021attentiongan}, DecentGAN \cite{xieunsupervised}, QS-GAN \cite{hu2022qs}, UNIT \cite{liu2017unsupervised}, SANTA \cite{xie2023unpaired}, VQ-I2I \cite{chen2022eccv}, UVCGAN \cite{torbunov2023uvcgan}, and StegoGAN \cite{wu2024stegogan}—or diffusion-based architectures such as UNSB \cite{kim2023unpaired}. By evaluating both paired and unpaired approaches using the same test dataset, we provide a comprehensive comparison of current state-of-the-art methods. Detailed differences between the chosen architectures are listed in Table \ref{tab:i2i_comparison}, and the mathematical formulations of both paired and unpaired methodologies are provided in \ref{suppsec:virtualStaining}.

\paragraph{Data preprocessing pipeline.} Accurate automated evaluation of virtual staining performance requires paired H\&E and IHC data, with the real IHC serving as ground truth. In this work, all evaluations are conducted on paired H\&E–IHC test sets, while the training data may be either paired or unpaired, depending on how each model’s data loader intakes  data \footnote{In the absence of paired test data, staining accuracy must be assessed either through expert pathological review or via performance on clinically relevant downstream tasks. However, manual review at the whole-slide image (WSI) level is highly labor-intensive and imposes significant time and cost burdens.}. We present our complete data processing pipeline (~\ref{fig:pipeline_full}), which includes four key components: (1) a \textbf{tissue piece extractor module} that identifies and registers corresponding H\&E and IHC tissue regions (~\ref{fig:TissueExtractionModule}); (2) a \textbf{patch extractor module} for balanced sampling of  stained and unstained regions to avoid data imbalances during training (~\ref{fig:PatchExtractor}); (3) a \textbf{virtual stainer module}, comprising multiple image-to-image (I2I) translation models (Supplementary Section~\ref{suppsec:virtualStaining}); and (4) an \textbf{image quality evaluation module}, providing an end-to-end framework to assess virtual IHC image quality (Supplementary Methods~\ref{suppsec:supp:methods}). The overall image generation and evaluation pipeline, along with the mathematical definitions of each evaluation metric, is detailed in the Methods (Table~\ref{tab:metrics_definition_table}) and Supplementary Methods (Section~\ref{suppsec:metrics}). Additional implementation details for all modules are included in the Supplementary Methods to facilitate reproducibility.

\begin{table*}[!htb]
    \centering
    \setlength{\tabcolsep}{10pt}
    \renewcommand{\arraystretch}{1.8}
    \begin{tabular}{p{4.6cm}|p{1.7cm}|c}
         \bf Metric Name&\bf Metric Type & \bf Metric Definition   \\
         \hline
         Fréchet Inception Distance (FID) & Distribution &  $FID(X,Y) = || \mu_X-\mu_Y||^2_2 + Tr(\sum_X +\sum_Y -2(\sum_X\sum_Y)^{\frac{1}{2}})$ \\
         Kernel Inception Distance (KID) &   & $KID(X,Y) = \mathbb{E}[ k(x,x')] + \mathbb{E}[ k(y,y')] -2 \mathbb{E}[ k(x,y)]$ \\
         Dist. Precision &  & $\text{Precision} = \frac{ \{y \in Y | y\in \mathcal{M}_r\} }{|Y|}$\\
         Dist. Recall &   &$ \text{Recall} = \frac{ \{x \in X | x\in \mathcal{M}_g\} }{|X|}$\\
         \hline 
        Signal to noise ratio(PSNR) & Texture  & $PSNR(\mathbb{I},\hat{\mathbb{I}})= 10. log_{10}\frac{255^2}{MSE}$ \\
        Structural Similarity(SSIM) &  & $SSIM(\mathbb{I},\hat{\mathbb{I}}) = \frac{(2\mu_\mathbb{I} \mu_{\hat{\mathbb{I}}} + c_1)(2\sigma_{\mathbb{I}\hat{\mathbb{I}}}+c_2)}{(\mu_\mathbb{I}^2+\mu^2_{\hat{\mathbb{I}}}+c_1)(\sigma^2_{\mathbb{I}} +\sigma^2_{\hat{\mathbb{I}}}+c_2)}$\\
         Mean Square Error (MSE) & &  $MSE(\mathbb{I},\hat{\mathbb{I}}) = \frac{1}{m\times n} \sum_m \sum_n ||\mathbb{I}(i,j)-\hat{\mathbb{I}}(i,j)||^2_2$\\
         \hline
         Dice Score & Segmentation  & $Dice = \frac{2*\mathbb{P}*\mathbb{GT}}{\mathbb{P}+\mathbb{{GT}}}$\\
         Intersection over union(IoU) &  & $IoU = \frac{\mathbb{P} \cap \mathbb{{GT}} }{\mathbb{P} \cup \mathbb{{GT}}}$\\
         Hausdorff Distance(HD) &  & $HD = \text{max}(h(\mathbb{P},\mathbb{{GT}}),h(\mathbb{{GT}},\mathbb{{P}})) $\\
         & &  $h(\mathbb{GT},\mathbb{{P}}) = \text{sup}_{a\in \mathbb{GT}} \text{inf}_{b\in \mathbb{{P}}} d(a,b)$ \\
         True Positive Rate (TPR)& & $\text{True Positive Rate} = \frac{\text{True Positives}}{(\text{True Positives}+\text{False Negatives})}$\\
         True Negative Rate(TNR) & & $\text{False Negative Rate} = \frac{\text{False Negatives}}{(\text{True Positives}+\text{False Negatives})}$\\
    \end{tabular}
    \caption{\textbf{Metrics Tables.} In the table above, $X$ denotes the latent space representations of the real IHC dataset, while $Y$ denotes those of the virtually generated IHC data, both extracted using a pretrained InceptionNet encoder. $\mathbb{I}$ represents a real IHC sample, and $\hat{\mathbb{I}}$ represents a corresponding virtual IHC sample. $\mathbb{GT}$ and $\mathbb{P}$ refer to the masks derived from these real and virtual IHC images, respectively.}
    \label{tab:metrics_definition_table}
\end{table*}

\subsection*{Evaluation Module}

Since one of the primary goals of virtually generated IHC images is to enable their use in IHC-based clinical workflows, it is crucial to establish pathologists’ trust in both their quality and accuracy. Achieving this requires not only high image fidelity—closely resembling real IHC stains—but also staining accuracy, ensuring that clinicians can intuitively assess and rely on the correctness of virtual IHC results. To address this, we present a comprehensive, automated, and quantitative evaluation framework for virtually stained pathology images. Although developed for virtual IHC, our approach is broadly applicable to other staining modalities, including chemical stains, immunohistochemistry, and immunofluorescence using antibody or oligonucleotide probes.

We use three categories of metrics to evaluate virtual staining performance(Table \ref{tab:metrics_definition_table}):
(1) Feature distribution-based metrics (FID, KID, Precision and Recall), which are widely used to assess the visual realism of generated images\cite{heusel2017gans,binkowski2018demystifying,li2023adaptive,kataria2024staindiffuser,kataria2025implicitstainer};
(2) Texture-based metrics (PSNR, SSIM, MSE), which measure the fidelity of tissue architecture;
(3) our proposed accuracy (segmentation-based) metrics (Dice, IoU, Hausdorff distance, and true positive/negative rates of stained cells), which we implement to determine IHC stain accuracy. This metric can also be used to determine hallucinations. We provide a more detailed description of these metrics below: 
\begin{itemize}
\item \textit{Distribution-based Metrics} such as FID \cite{heusel2017gans}  and KID \cite{binkowski2018demystifying},  precision and recall \cite{kynkaanniemi2019improved}, measure the similarity in high dimentational feature(or latent space representation) distributions between real and virtual IHC images. These metrics rely on latent space representations obtained via an image encoder and do not require paired datasets, making them applicable to both paired and unpaired test datasets. While they do not directly assess staining accuracy, they serve as useful indicators of the similarity of color and structure feature distributions between real and generated images, helping to evaluate overall image quality. FID and KID measure the distance between real and generated data distributions in latent feature space, with FID assuming Gaussian-distributed embeddings and KID relying on a polynomial kernel. In contrast, \textit{precision} and \textit{recall} capture complementary aspects of generative performance: precision reflects the diversity of generated samples, while recall quantifies their coverage relative to the real data distribution. These metrics are estimated using a k-nearest neighbors approach to approximate the underlying data manifolds. Mathematical formulation of these metrics are defined in supplementary section \ref{suppsec:metrics:distribution}. 
\item \textit{Texture Based Metrics (PSNR, SSIM, MSE)}: These metrics rely on pixel-level comparisons between real and generated images. While such metrics can offer useful insights into tissue and cell restruction, they are insufficient for comprehensively evaluating the accuracy and utility of virtual staining. 

\item \textit{Staining Accuracy Evaluations}. Automated, objective and quantitative evaluation of staining accuracy depends on the availability of reliable, pixel-level ground truth. Therefore, evaluation of stain accuracy requires a test set of H\&E and IHC patches that are paired with pixel-level precision. If H\&E and IHC are on adjacent tissue sections, an automated and scalable evaluation of virtual staining accuracy is not possible because the cells in adjacent tissue sections are not the same. 

We propose two methods for measuring and communicating staining accuracy, both based on segmenting the brown IHC regions in real and virtually stained images. This segmentation-based evaluation treats the brown-stained areas(or IHC positive pixels) as binary masks, allowing the use of metrics such as Dice score, Intersection over Union (IoU),  Hausdorff distance, true positive rate, and true negative rate. These metrics are computed only on test patches that contain positive (brown-stained) pixels. Notably, the reliability of this evaluation depends on the accuracy of the segmentation masks derived from brown pixel detection. Therefore, we use two different approaches for IHC brown mask generation:
\begin{itemize}
    \item We convert the RGB image into the HED color space and manually threshold the DAB channel to generate a binary mask \cite{macenko2009method}. Importantly, this RGB decomposition must be performed on the full tissue image to determine an appropriate threshold; applying it to small patches often yields inaccurate results due to limited color context. Moreover, when a large proportion of cells are DAB-stained, the decomposition can become noisy, as the algorithm lacks sufficient unstained background to reliably isolate the DAB signal.
    \item To overcome the limitations of the thresholding method, we train a U-Net–based segmentation model on annotations obtained in the HED color space \cite{kataria2023automating}, using a low learning rate to ensure robust learning of brown-stained pixels and regions (section \ref{suppsec:metrics:modelsegmentation}). After training, the model weights are frozen and the fixed model is used to generate segmentation masks for both real and virtual IHC images. We then compute IoU and Dice scores between the masks from real and virtual images. Since the segmentation model remains fixed, the results offer an objective measure of how accurately virtual staining and true staining overlap. We manually confirm that the segmentation generated by the trained model accurately segments the brown region in real and virtual IHC. 
\end{itemize}

    \item \textit{Manual quality evaluations on WSIs.} We evaluate the effectiveness of virtual staining by incorporating virtually stained IHC images into our existing automated annotation pipeline that we previously designed for real IHC images (\textit{Kataria et al. 2023} \cite{kataria2023automating}). We transfer the IHC masks to the H\&E images and train segmentation models on IHC-derived gland annotations. The performance of the segmentation models is compared to the model trained on transferred gland outlines from real IHC images. 
\end{itemize}

To our knowledge, this is the first evaluation pipeline combining traditional image generation metrics, segmentation measures, and WSI-level quantitative analysis. Segmentation metrics derived from brown IHC masks provide a direct assessment of staining accuracy. Mathematical definitions for all metrics are provided in Supplementary Methods Section~\ref{suppsec:metrics}.

\subsection*{Statistical Analysis and Latent Space Visualization}

\textit{Calculation of p-values for comparing metrics obtained from different models}: Null hypothesis testing in statistics assumes that data samples are drawn independently. To reduce the technical variability, models are trained using the same tiles, parameters, epochs and computer hardware. The test set is the same for all the models.  We apply a standard t-test to assess statistical significance between groups. We report p-values separately for paired and unpaired models to evaluate significance within each training framework — specifically, to determine whether models trained under the same data pairing conditions exhibit statistically significant differences in performance. P-value ($p<0.05$) is considered statistically significant in our experiments. 

\noindent \textit{UMAP visualizations.} We used UMAP \cite{mcinnes2018umap} to distinguish between samples generated by Pix2Pix \cite{isola2017image} and CycleGAN \cite{zhu2017unpaired}, representing two major families of image translation models. Both H\&E and IHC stains contain hematoxylin (which highlights nuclei), but only IHC includes the DAB stain (brown coloration). To visualize the similarity between real and generated IHC images, we focus on the hematoxylin and DAB channels. Since most image encoders operate on RGB inputs, we first converted the RGB images to HED space. To isolate the hematoxylin channel, we set the eosin and DAB components to zero and converted back to RGB. A similar procedure was followed to isolate and visualize the DAB channel.

\subsection*{Data Collection and Implementation Details}

\paragraph{Data collection.} The use case for our virtual staining pipeline consists of a cohort of surveillance biopsies from individuals diagnosed with inflammatory bowel disease. The dataset comprises H\&E-stained and CDX2 restained whole slide images (WSIs) from five patients with ulcerative colitis. The cohort consists of 92 tissue pieces, ranging from 16 to 24 tissues per slide. The staining process of the glass slides is described in Kataria et. al. \cite{kataria2023automating}. Briefly, formalin-fixed, paraffin embedded (FFPE) tissue blocks were retrieved from the archive of the pathology department at the University of Utah under an IRB approved protocol (IRB \#00091019). Glass slides were first stained with hematoxylin and eosin (H\&E) using an automated clinical staining system, and scanned at 40x magnification (0.25 $\mu$m/pixel resolution) using an Aperio AT2 slide scanner. Following scanning, coverslips were removed, and the slides were restained with the CDX2 antibody (clone number EP25) via immunohistochemistry (IHC) using the Leica Bond III autostainer. Heat-induced epitope retrieval before antibody incubation effectively decolorized the H\&E stain, eliminating the need for manual destaining. The IHC-stained slides were then scanned using the Aperio AT2 at 40x, and each resulting digital IHC image was registered to its corresponding H\&E image.

\paragraph{Implementation Details.} Training patches were extracted from 70 WSIs each containing one tissue piece, reserving the rest of the tissue pieces for testing. Consistent with conventional training protocols, the models were trained for 200 epochs with default parameters. All models were trained on NVIDIA A100-40GB GPUs. All the evaluations on the virtually generated results are done inline with the evaluation pipeline proposed above, where we report distribution metrics, texture metrics and staining accuracy metrics.

\section*{Data Availability Statement}
Data from the University of Utah will be available after the execution of a data-sharing agreement. Contact one of the corresponding authors of the paper for further details.
\section*{Code Availability Statement}
The code for training all image translation models is already publicly available, with links provided in the supplementary information section. Similarly, the evaluation code is taken from existing public libraries. A complete list of these repositories is included in Supplementary Section \ref{ssup:code}. The scripts used for evaluation are added in a public github repo released at \href{https://github.com/tushaarkataria/Virtual-Staining-Evaluation}{Virtual-Staining-Evaluation}.

\bibliography{sample}


\section*{Acknowledgements}
We thank the Department of Pathology and the Kahlert School of Computing at the University of Utah for their support of this project. The support and resources from the Center for High Performance Computing at the University of Utah are gratefully acknowledged. The computational resources used were partially funded by the NIH Shared Instrumentation Grant 1S10OD021644-01A1.

\section*{Author contributions statement}
Training of algorithms and code (TK, SD), conceptual framework (TK, SD, BK, SE), Statistical Evaluations (TK, BB, BK), and manual annotations (MB, JJ, BK). (TK, SD, BB, BK, SE) wrote the manuscript text and figures. All authors reviewed the manuscript.

\section*{Competing Interests}
The authors have no conflicts of interest to declare that are relevant to the content of this article.

\section*{Ethical Statement}
All experimental protocols were approved by the Institutional Review Board (IRB) at the University of Utah under IRB\_00140202 and IRB\_00057287. All the experiments in the manuscript followed the guidelines of the IRB protocols. The study was approved by the IRB under a waiver of consent since all HIPAA-sensitive data fields were removed prior to the use of patient slides. No demographic or clinical information from study participants was used for data analysis, and the link to the medical record was destroyed before the images were processed.

\clearpage
\onecolumn
\section*{Supplementary Material}
\setcounter{figure}{0}
\renewcommand{\thefigure}{SF\arabic{figure}}
\setcounter{section}{0}
\renewcommand{\thesection}{S\arabic{section}}
\setcounter{table}{0}
\renewcommand{\thetable}{ST\arabic{table}}
\setcounter{page}{1}

\vspace{1em}
\textbf{Supplementary Figures} \\
\begin{table}[!thb]
    \centering
    \begin{tabular}{p{1.0cm}@{} p{13.6cm}@{} c}
    \textit{}  & & \textit{Page}  \\
         SF1 & \textbf{Examples of virtual CDX2 IHC WSIs..} & 2 \\
         SF2 &  \textbf{Examples of virtual CDX2 IHC WSIs..} & 3 \\
         SF3 &  \textbf{FID comparison with pathology-pretrained encoder vs ImageNet encoder.} & 4 \\
         SF4 &  \textbf{Comparison of UMAP distances and vector distances of paired real and virtual image tiles.} & 4 \\
         SF5 &  \textbf{Violin plots for texture metrics.} & 5 \\
         SF6 &  \textbf{Stain accuracy masks from DAB thresholding and model segmentation.} &  \\
         SF7 &  \textbf{Voilin plots of patch-wise stain accuracy values..} & 7 \\
         SF8 &  \textbf{Tile-wise correlation heat-maps of texture and segmentation metrics. } & 8 \\
         SF9 &  \textbf{Visualization of model-wise and tile-wise correlations.} & 8 \\
         SF10 &  \textbf{Blocky artifacts in WSI prediction.} & 9 \\
         SF11 &  \textbf{Virtual Staining Pipeline.} & 12 \\
         SF12 &  \textbf{Tissue Extraction Module.} & 13 \\
         SF13 &  \textbf{Patch Extractor Module.} & 14 \\
    \end{tabular}
\end{table}

\textbf{Supplementary Tables} \\
\begin{table}[!thb]
    \centering
    \begin{tabular}{p{1.0cm}@{} p{13.6cm}@{} c}
 \textit{}  &  & \textit{Page}  \\
         ST1 & \textbf{Feature distribution metrics using pathology specific domain encoders.} & 10 \\
         ST2 &  \textbf{Image quality as scored by a pathologist.} & 10 \\
         ST3 &  \textbf{p-values from t-test comparing texture metrics  in image tiles with perfect versus imperfect manual image quality
evaluation scores.} & 10 \\
         ST4 &  \textbf{Table of acronyms.} & 11 \\
         ST5 &  \textbf{Model name, year and unique attributes.} & 17 \\
    \end{tabular}
\end{table}

\textbf{Supplementary Methods}

\begin{table}[!thb]
    \centering
    \begin{tabular}{p{1.0cm}@{} p{13.6cm}@{} c}
 \textit{}  &  & \textit{Page}  \\
         S 2.1 & \textbf{Tissue Registration and DAB Mask in the Extraction Module.} & 12 \\
         S 2.2 &  \textbf{Patch Extractor Module.} & 14 \\
         S 3 &  \textbf{Virtual Staining Training Models.} & 10 \\
         S 4 &  \textbf{Evaluation Metrics Details.} & 16 \\
         S 4.1 &  \textbf{Distribution Metrics.} & 16 \\
         S 4.2 &  \textbf{Texture Metrics Equations.} & 17 \\
         S 4.3 &  \textbf{Segmentation Model Training for Segmentation-Based Metrics.} & 18 \\
         S 4.4 &  \textbf{Segmentation Metrics.} & 18 \\
         S 4.5 &  \textbf{Code Repositories Used.} & 18 \\
    \end{tabular}
\end{table}

\newpage

\section{Supplementary Figures and Tables}\label{ssup:figures_tables}

\begin{figure*}[!htb]
    \centering
    \includegraphics[trim={0.0cm 20.3cm 15.0cm 0.0cm}, clip=true,width=1.0\linewidth]{WSI-Full-Predictions.pdf}
    \caption{\textbf{Examples of virtual CDX2 IHC WSIs.} WSI predictions across image-to-image translation models listed above the image. The same H\&E slide was used by the trained models to generate the WSI IHC slide. The higher magnification insert shows nuclear CDX2 staining specific to glandular epithelial cells.}
    \label{fig:visual_wsi_1}
\end{figure*}

\begin{figure*}[!htb]
    \centering
    \includegraphics[trim={0.0cm 20.3cm 15.0cm 0.0cm}, clip=true,width=1.0\linewidth]{WSI-Prediction-Examples.pdf}
    \caption{\textbf{Examples of virtual CDX2 IHC WSIs.} WSI predictions across image-to-image translation models listed above the image. The same H\&E slide was used by the trained models to generate the WSI IHC slide. The higher magnification insert shows nuclear CDX2 staining specific to glandular epithelial cells.}
    \label{fig:visual_wsi_2}
\end{figure*}

\begin{figure}[!thb]
    \centering
    \includegraphics[width=0.9\linewidth]{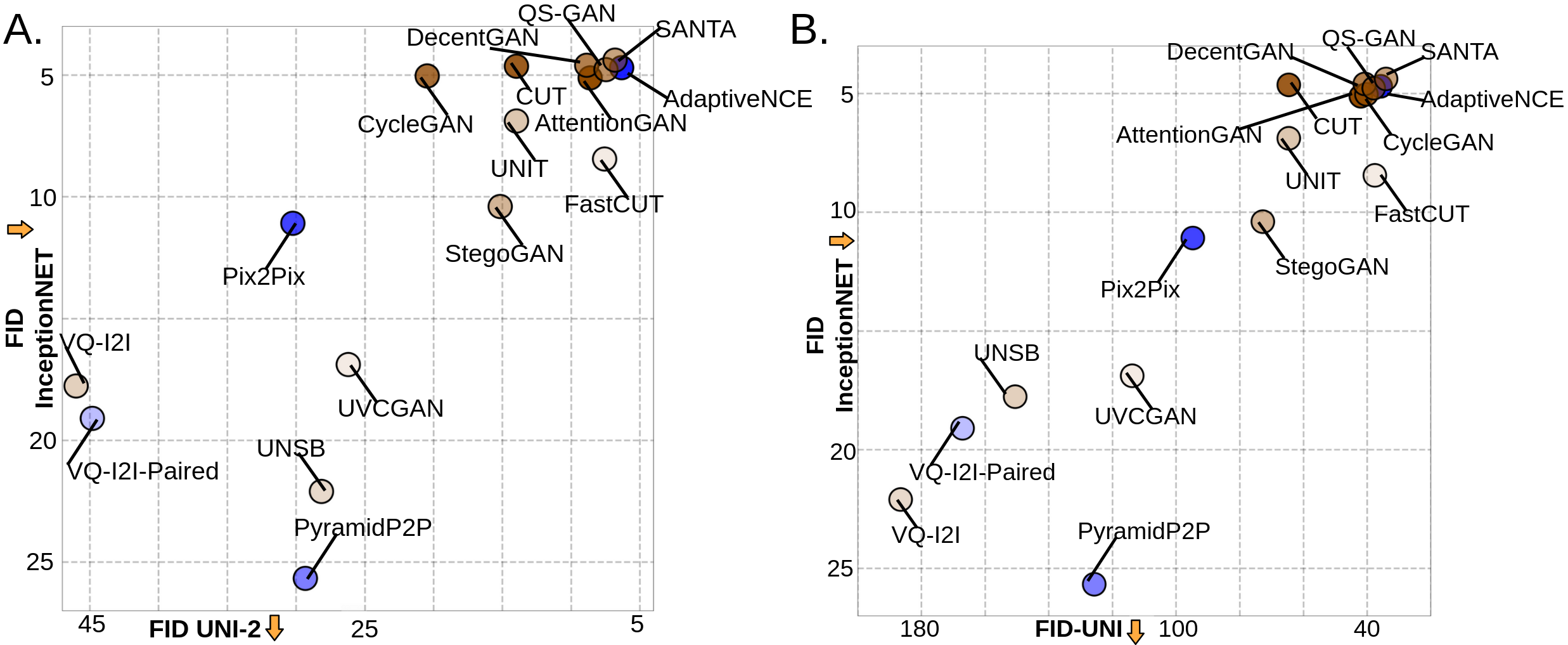}
    \caption{\textbf{FID comparison with pathology-pretrained encoder vs ImageNet encoder.} Scatter plots comparing FID scores obtained using the standard Inception-Net versus those computed with pathology-specific pretrained models, UNI and UNI-2.}
    \label{fig:fid_fid_uni}
\end{figure}
\begin{figure}
    \centering
    \includegraphics[width=1.0\linewidth]{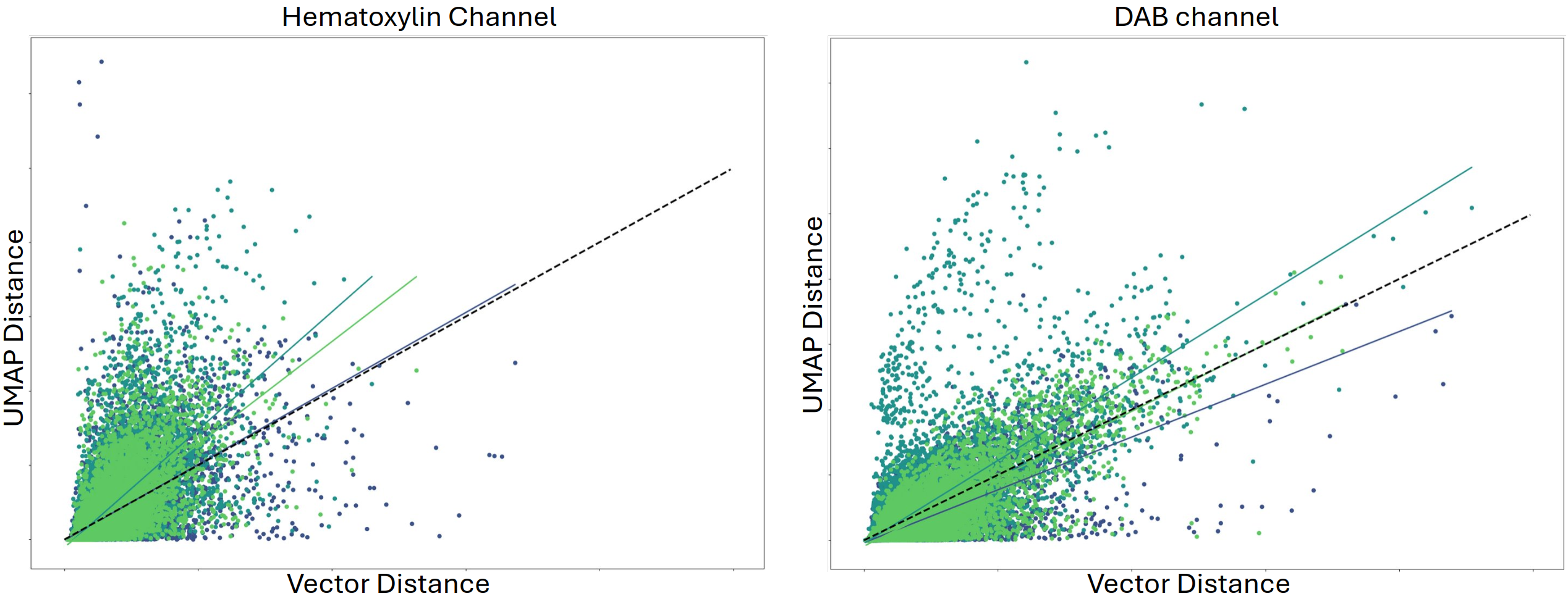}
    \caption{\textbf{Comparison of UMAP distances and vector distances of paired real and virtual image tiles.} Scatter plots of 2D linear UMAP projection distance and high dimensional embedding distance between pairs of real and virtual image tiles. Hematoxylin and DAB channel were separated prior to input into the I2I translation model. \textcolor{blue}{Navy dots} correspond to real images, \textcolor{teal}{teal dots} correspond to Pix2Pix virtual images and \textcolor{green}{green dots} correspond to CycleGAN virtual images.}
    \label{fig:UMAP-L2}
\end{figure}

\begin{figure*}[!htb]
    \centering
    \includegraphics[width=1.0\linewidth]{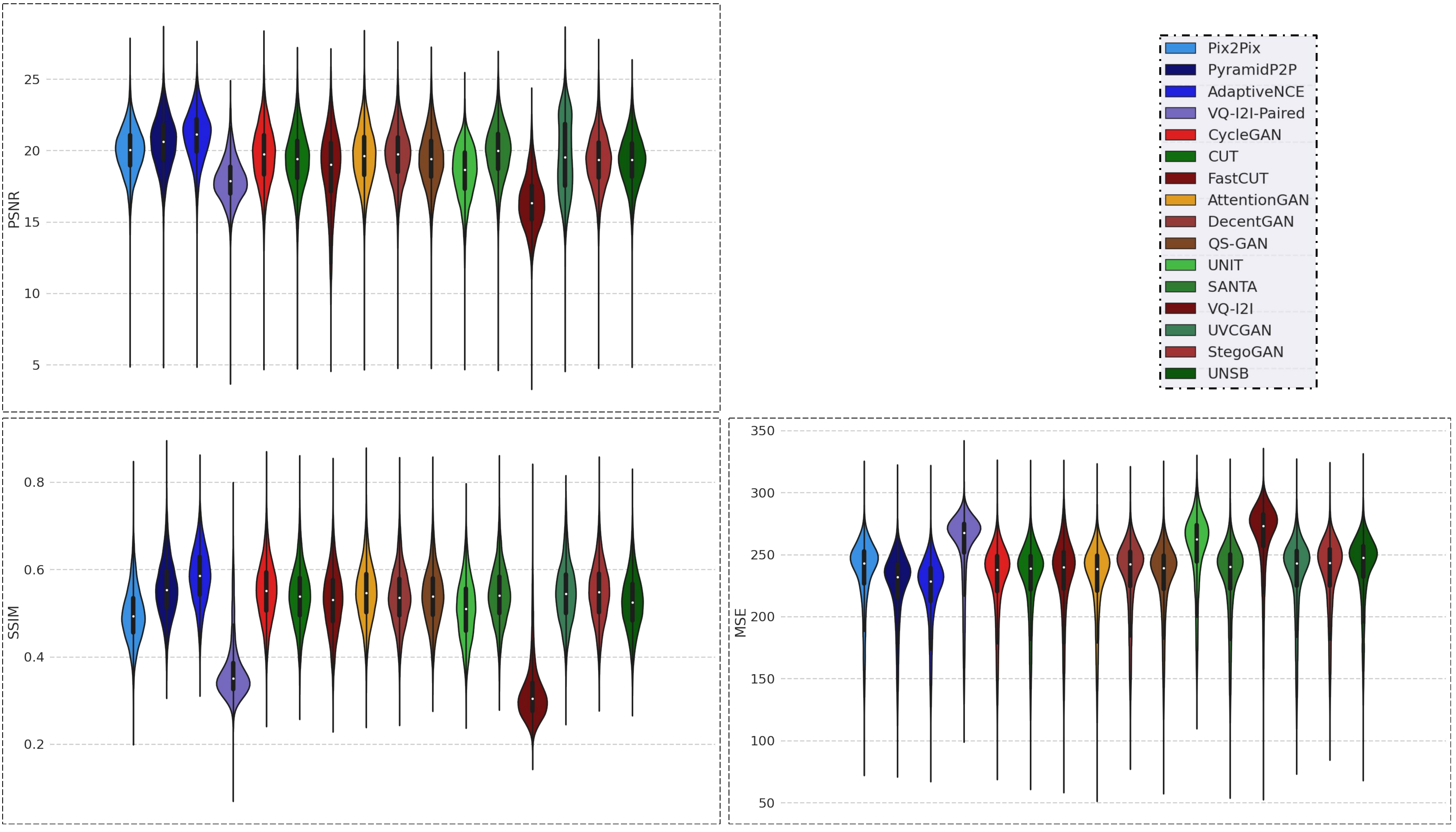}
    \caption{\textbf{Violin plots of texture metrics.} The Violin plots show the distribution of patch-wise texture metrics for all I2I translation models.}
    \label{fig:texture_voilin}
\end{figure*}

\begin{figure}[!htb]
    \centering
    \includegraphics[width=0.75\linewidth]{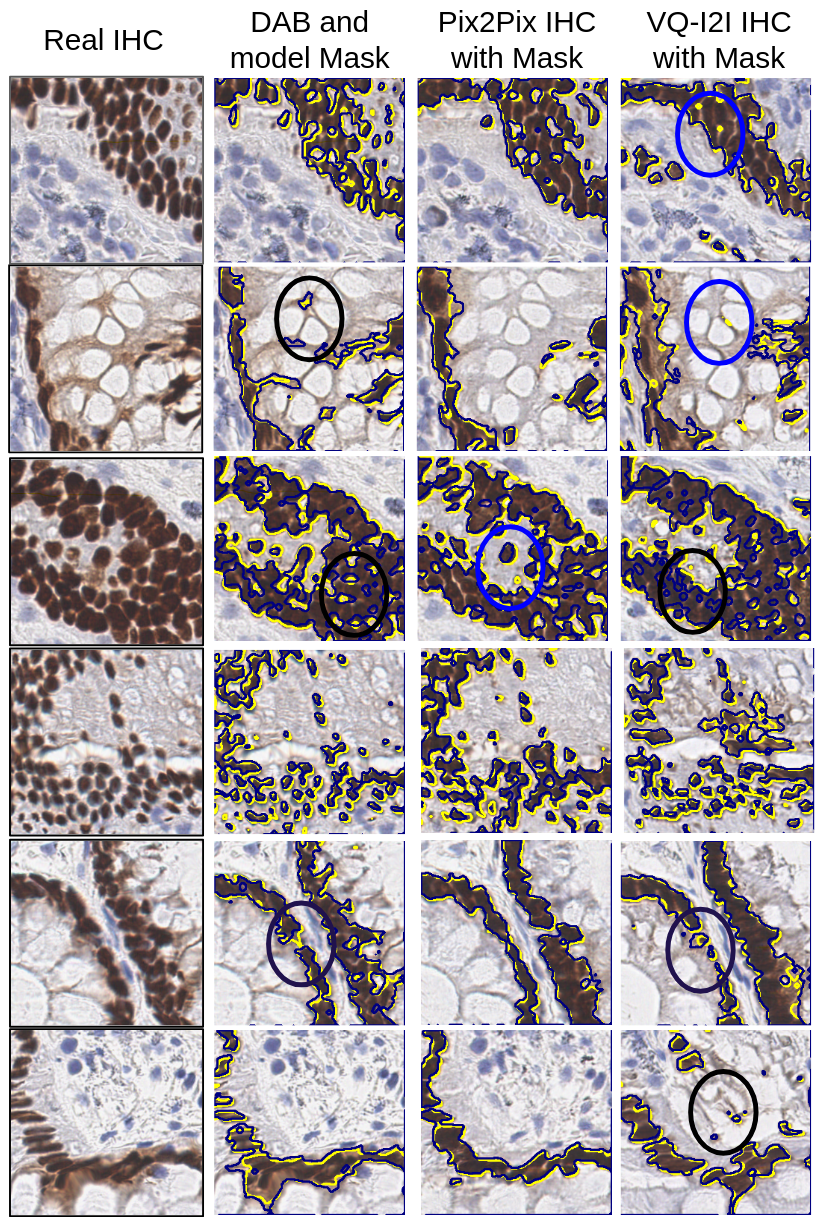}
    \caption{\textbf{Stain accuracy masks from DAB thresholding and model segmentation}. IHC stain pixel masks obtained using DAB mask thresholding (\textit{navy line}) and model segmentation (yellow line). The model was pretrained.  \textcolor{blue}{\textbf{Blue}} circles indicate places where model segmentation gives wrong masks, \textbf{Black} circles indicate places where DAB thresholding gives wrong masks. }
    \label{fig:dab_model_segmentation}
\end{figure}
\begin{figure*}[!htb]
    \centering
    \includegraphics[trim={0.0cm 8.0cm 0cm 0.0cm}, clip=true, width=0.95\linewidth]{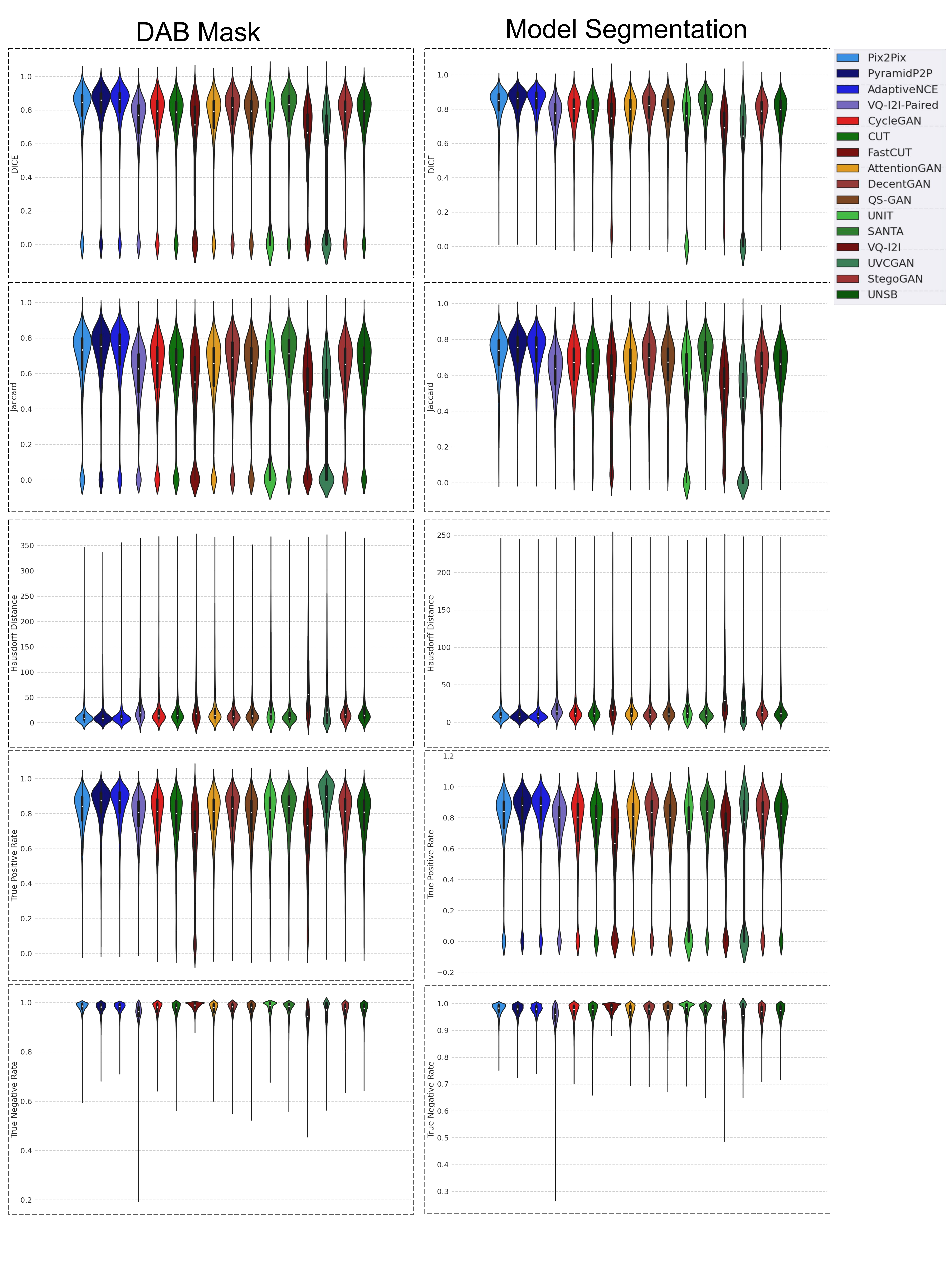}
    \caption{\textbf{Voilin plots of patch-wise stain accuracy values.} The Figure shows the violin plots for all segmentation metrics, for both DAB-based mask generation and model-based mask generation.}
    \label{fig:model_segmentation_voilin}
\end{figure*}

\begin{figure*}[!htb]
    \centering
    \includegraphics[width=1.0\linewidth]{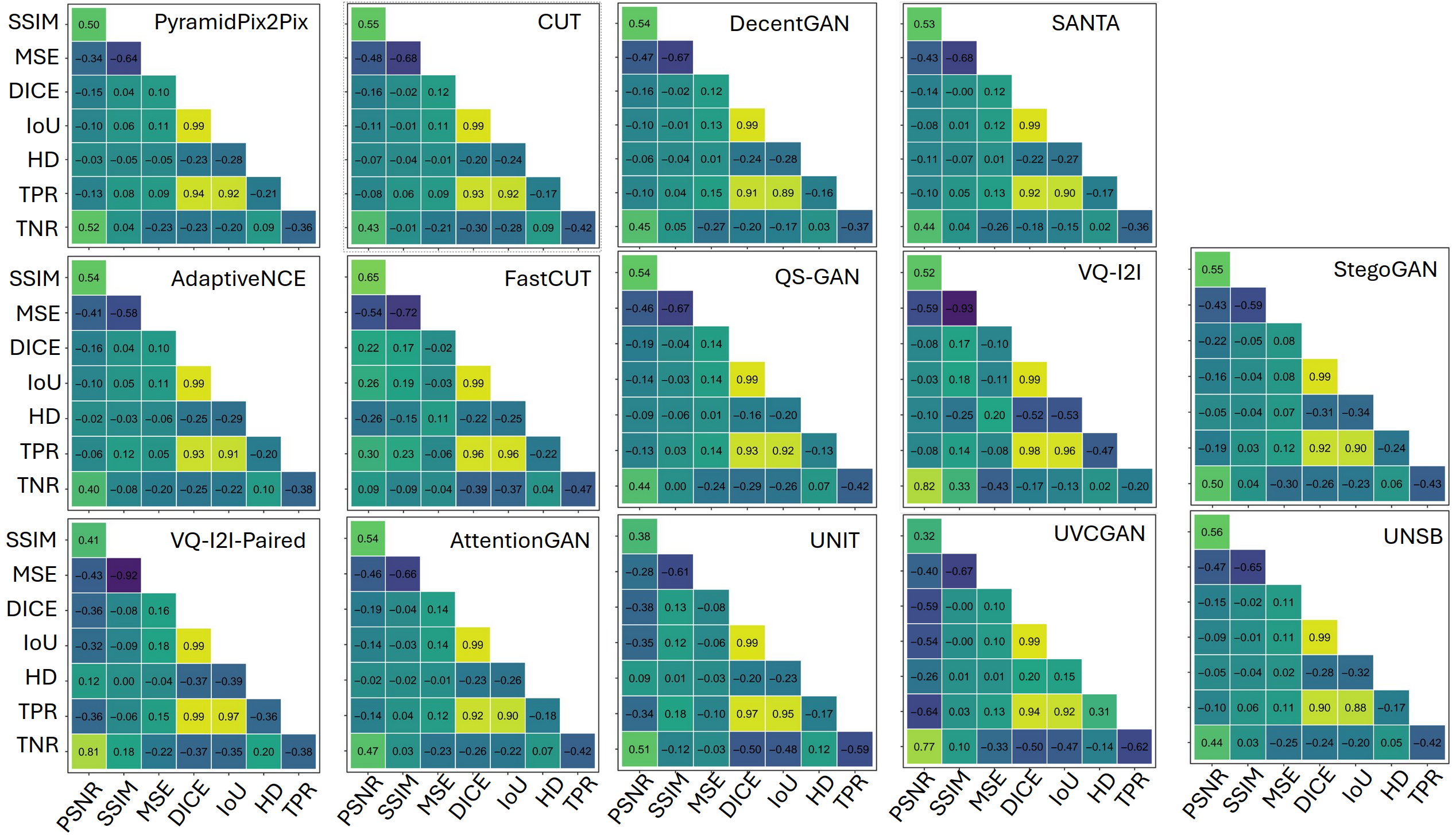}
    \caption{\textbf{Tile-wise correlation heat-maps of texture and segmentation metrics.} For each model pair-wise correlations are shown between texture and segmentation metrics. The model name is listed in the right upper corner.}
    \label{fig:correlation_maps}
\end{figure*}

\begin{figure*}[!htb]
    \centering
    \includegraphics[width=1.0\linewidth]{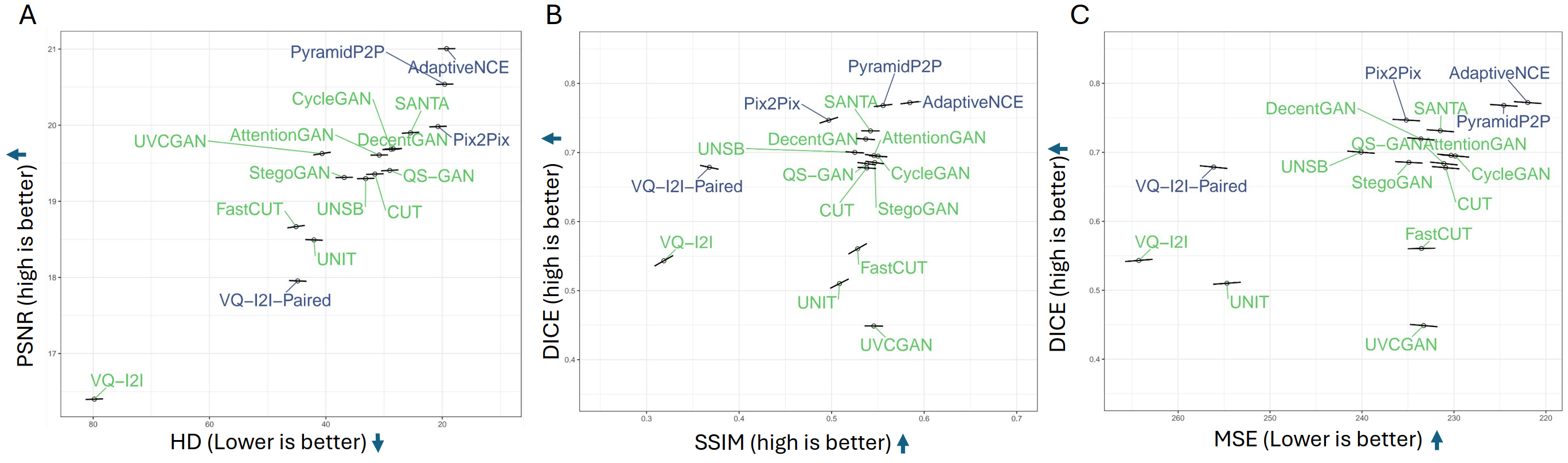}
    \caption{\textbf{Visualization of model-wise and tile-wise correlations.}  The average metrics values are marked by a green dot in the middle of a black line. The slopoe of the black line indicates the tile-wise relationship between metrics. The values of tile-wise relationships are shown in SF8.   \textbf{A.} Scatter plot between PSNR and Hausdorff Distance. \textbf{B.} Scatter plot DICE and SSIM. \textbf{C.} Scatter plot between DICE and MSE.}
    \label{fig:other_metric_correlations}
\end{figure*}

\begin{figure*}[!htb]
    \centering
    \includegraphics[trim={0.0cm 49.0cm 30.0cm 0.0cm}, clip=true,width=1.0\linewidth]{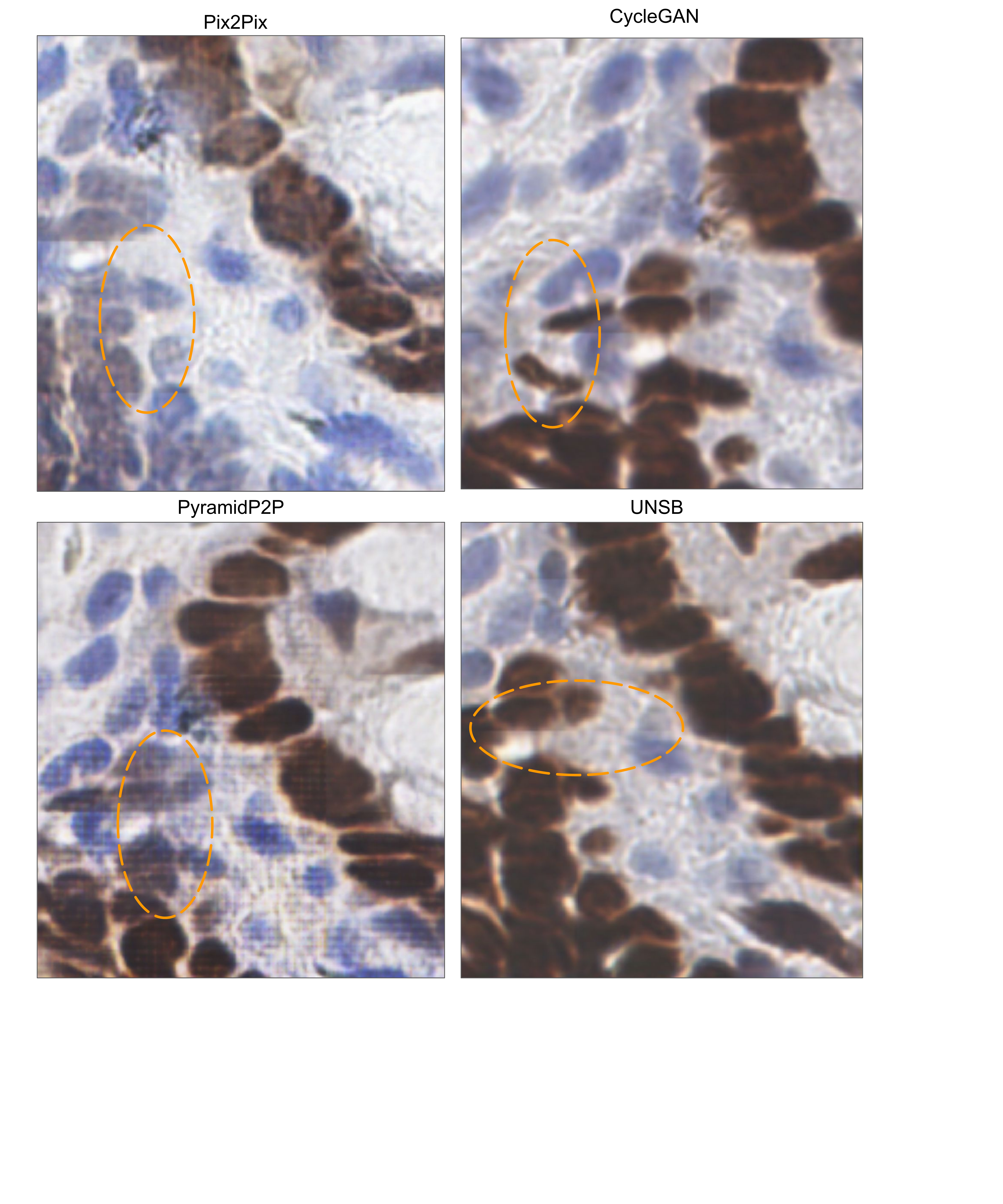}
    \caption{\textbf{Artifacts at tile boundaries in WSI prediction.} Whole-slide image (WSI) predictions are generated using a sliding window with a 192-pixel overlap per patch. Despite this strategy, smoothing  artifacts remain visible in the final reconstructed images. Orange circles highlight the actual horizontal and vertical lines, which are affected by window prediction.}
    \label{fig:block_artifacts}
\end{figure*}
\clearpage
\begin{table*}[!htb]
    \centering
\begin{tabular}{c||c|c|c|c||c|c|c|c}
     &  \multicolumn{4}{c||}{\bf UNI Encoder\cite{chen2024uni}} & \multicolumn{4}{c}{\bf UNI-2-h Encoder\cite{chen2024uni}} \\
     \hline
Model  & FID $\downarrow$ & KID $\downarrow$ & Dist. Prec. $\uparrow$ & Dist. Recall $\uparrow$ & FID $\downarrow$ & KID $\downarrow$ & Dist. Prec. $\uparrow$ & Dist. Recall $\uparrow$\\
    \hline
    \hline
Pix2Pix\cite{isola2017image}       &  94.74	& 0.3025	& 0.7498	& 0.5469	& 30.23	 & 0.0376	& 0.5756	& 0.4683\\
PyramidPix2Pix \cite{liu2022bci}   &  125.7	& 0.5297	& 0.5467	& 0.5748	& 29.32	 & 0.0402	& 0.4925	& 0.5532\\
AdaptiveNCE \cite{li2023adaptive}  & \bf 35.73 &  \bf 0.0803	& \bf 0.9534	& \bf 0.8893	& \bf 6.32	& \bf 0.0048	& \bf 0.9335	& \bf 0.9072\\
VQ-I2I-Paired  \cite{chen2022eccv} & 167.06 & 0.6258	& 0.2641	& 0.0646	& 44.81	 & 0.048	& 0.2385	& 0.0348\\
    \hline 
    \hline 
CycleGAN \cite{zhu2017unpaired}  &  40.2	& 0.0685	& 0.8489	& 0.83	& 20.46	& 0.0081	& 0.8281	& 0.842 \\
CUT       \cite{park2020contrastive} &  64.6	& 0.1322 &	0.7552& 	0.7152 &	13.97	& 0.0122	& 0.75	&  0.7431\\
FastCUT    \cite{park2020contrastive} & 37.56 & 0.0637	& 0.829	& 0.8111 & 7.56 &	0.0049	& 0.8158	& 0.8264\\
Attention GAN  \cite{tang2021attentiongan}  &  41.95 &	0.0834	& 0.8573 & 0.8182	& 8.61 &	0.0069	& 0.8338	& 0.8407\\
Decent GAN    \cite{xieunsupervised}    &  40.74	& 0.0793	& 0.86	& 0.7889	& 8.86	& 0.006	& 0.8222	& 0.79  \\
QS-GAN \cite{hu2022qs}         & 37.92	& 0.0665	& 0.8324	& 0.8155	& 7.459	& 0.005	& 0.8108	& 0.8313\\
UNIT \cite{liu2017unsupervised} & 64.6	& 0.1322 & 0.7552	& 0.7152	& 13.97 & 0.0122 & 0.75	& 0.7431 \\
SANTA \cite{xie2023unpaired} &  \bf 34.05 &	\bf 0.0639 & \bf 0.8788 &	\bf 0.8518 & \bf 6.801	& \bf 0.0041 & \bf 0.8787 & \bf 0.8364  \\
VQ-I2I \cite{chen2022eccv} & 150.54	& 0.4497 & 0.1748	& 0.0843	& 45.992	& 0.0547	& 0.1312	& 0.0506\\
UVCGAN    \cite{torbunov2023uvcgan} &  113.78 & 0.4029	& 0.472	& 0.4758	& 26.2	& 0.03	& 0.4406	& 0.4999 \\
StegoGAN  \cite{wu2024stegogan}  &  72.75 &	0.2308 & 0.7211	& 0.7213 & 15.15 & 0.0154 & 0.7383	& 0.6945\\
UNSB      \cite{kim2023unpaired}  &  186.51 & 0.8427 & 0.2813 & 0.2683	& 28.15	& 0.0332	& 0.441	& 0.3882 \\
    \end{tabular}
    \caption{\textbf{Evaluation of virtual images using standard feature distribution metrics using pathology specific domain encoders}. The similarity between real and virtual images is assessed using Fréchet Inception Distance (FID) \cite{heusel2017gans}, Kernel Inception Distance (KID) \cite{binkowski2018demystifying}, and feature distribution precision and recall \cite{kynkaanniemi2019improved} evaluated over the full dataset. Distribution precision and recall quantify the diversity and coverage of the feature distribution of virtual IHC images relative to real images, while FID and KID measure the distance between the two feature distributions. 
    }
    \label{tab:pathology_domain_distribution_metrics}
\end{table*}

\begin{table*}[!htb]
    \centering
    \begin{tabular}{p{2.8cm}@{}|p{1.6cm}@{}|p{11.6cm}@{}}
 \rowcolor{gray}   \bf Category & \bf Label & \bf Definition \\
    \hline
      \rowcolor{lightgray}  Cell Morphology & Good/Bad &  The virtual image accurately depicts the size, shape and texture of individual cells and nuclei\\
      \rowcolor{white}  Image Blurring & Good/Bad & The tissue structure and individual cells and nuclei are distinct with sharp edges \\
       \rowcolor{lightgray} Hallucination &  Yes/No & The generated virtual image does not introduce cells that are absent in the corresponding reference H\&E image. \\
       \rowcolor{white} Color fidelity & Good/Bad & The coloration of the image is realistic, i.e. all negative nuclei are colored in blue versus for example in red.\\
       \rowcolor{lightgray} Tissue Architecture & Good/Bad & Multicellular structures in tissues such as glands, vessels, nerves etc. are realistic\\
    \end{tabular}
    \caption{\textbf{Characteristics of image quality as scored by a pathologist.} }
    \label{tab:manual_annotation_descritpion}
\end{table*}

\begin{table}[!htb]
\centering
\begin{tabular}{l|ccc}
\toprule
 &  \multicolumn{3}{c}{\bf p-values}  \\
\textbf{Method} & \textbf{PSNR} & \textbf{SSIM} & \textbf{MSE} \\
\midrule
Pix2Pix          & 0.5458 & 0.2975 & 0.9562 \\
CycleGAN         & \textcolor{blue}{0.0015} & 0.0111 & 0.0634 \\
AdaptiveNCE      & 0.3156 & 0.6816 & 0.5963 \\
PyramidPix2Pix   & 0.6261 & 0.0410 & 0.1540 \\
UNSB             & \textcolor{blue}{0.0002} & \textcolor{blue}{0.0017} & 0.0134 \\
\bottomrule
\end{tabular}

\caption{\textbf{Comparison of texture metrics between groups of tiles with perfect versus imperfect manual quality evaluation scores.} Tiles were individually scored by a pathologist using 5 scoring parameters. For each model (Pix2Pix, CycleGAN, ADaptiveNCE, PyramidPix2Pix and UNSB) tiles with only perfect scores were separated from tiles with at least one imperfect score. The texture metrics, PSNR, SSIM and MSE were compared between the perfect and imperfect group using a t-test. The  p-values for each model and texture metric are listed in the table.}
\label{tab:texture_metrics_p_values}
\end{table}

\begin{table}[!htb]
    \centering
    \begin{tabular}{c|c}
    Acronyms & Full Form \\
        \hline
        H\&E &  Hematoxylin and Eosin Stain\\
        IHC &  immunohistochemical stain\\
        DAB & diaminobenzidine \\
        WSI & Whole Slide images \\
        CD3 & Cluster of Differentiation 3 \\
        CDX2 & caudal type homeobox 2 \\
        HER2 & Human Epidermal Growth Factor Receptor 2 \\
        IBD & Inflammatory Bowel Disease \\
        ER & Estrogen receptor \\
        Ki67 & Antigen Kiel 67 \\
        GAN & Generative Adverserial Networks \\
        PSNR & Signal to Noise Ratio \\
        MSE & Mean Square Error \\
        SSIM & Structural Similarity Index \\
        FID & Fréchet Inception Distance \\
        KID & Kernel Inception Distance \\
        RGB & Red, Green Blue Color Space \\
        UNI UNI-2 & Pretrained Histopathology Encoders\cite{chen2024uni}\\
        IoU & intersection over Union\\
        HD & Hausdorff Distance \\
        TPR & True Positive Rate \\
        TNR & True Negative Rate \\
        
    \end{tabular}
    \caption{\textbf{Table of acronyms}.}
    \label{tab:acronyms}
\end{table}

\clearpage

\section{Supplementary Methods} \label{suppsec:supp:methods}
The complete proposed processing pipeline is illustrated in Figure \ref{fig:pipeline_full}. It consists of three main components: (a) a preprocessing module, which involves isolating tissue sections, registering of H\&E and IHC, and sampling patches for training; (b) the virtual staining module, which employs image-to-image translation models to generate virtual IHC stains; and (c) the evaluation module, which defines the performance assessment procedures described in the main paper. In the following sections, we describe the remaining sub-modules in detail to support reproducibility within the research community.

\subsection{Tissue Registration and DAB Mask in the Extraction Module}\label{suppsec:tissueregistation}
The tissue extraction module includes the following steps: background removal, registration, and the generation of binary masks from the DAB stain and transfer of the masks from IHC to H\&E image. The H\&E and IHC registration steps differ when the slide displays multiple small biopsies versus a single large piece of tissue. The registration of small pieces is described in \textit{Kataria et al. 2023} \cite{kataria2023automating} and focuses on isolating tissue pieces for registration by removing the white background. Pixel-level registration uses the ANTsPy library \footnote{https://antspy.readthedocs.io/en/latest/registration.html}. For \textit{unpaired samples}, only background removal is necessary. 

\begin{figure*}[!htb]
    \centering
    \includegraphics[width=1.0\linewidth]{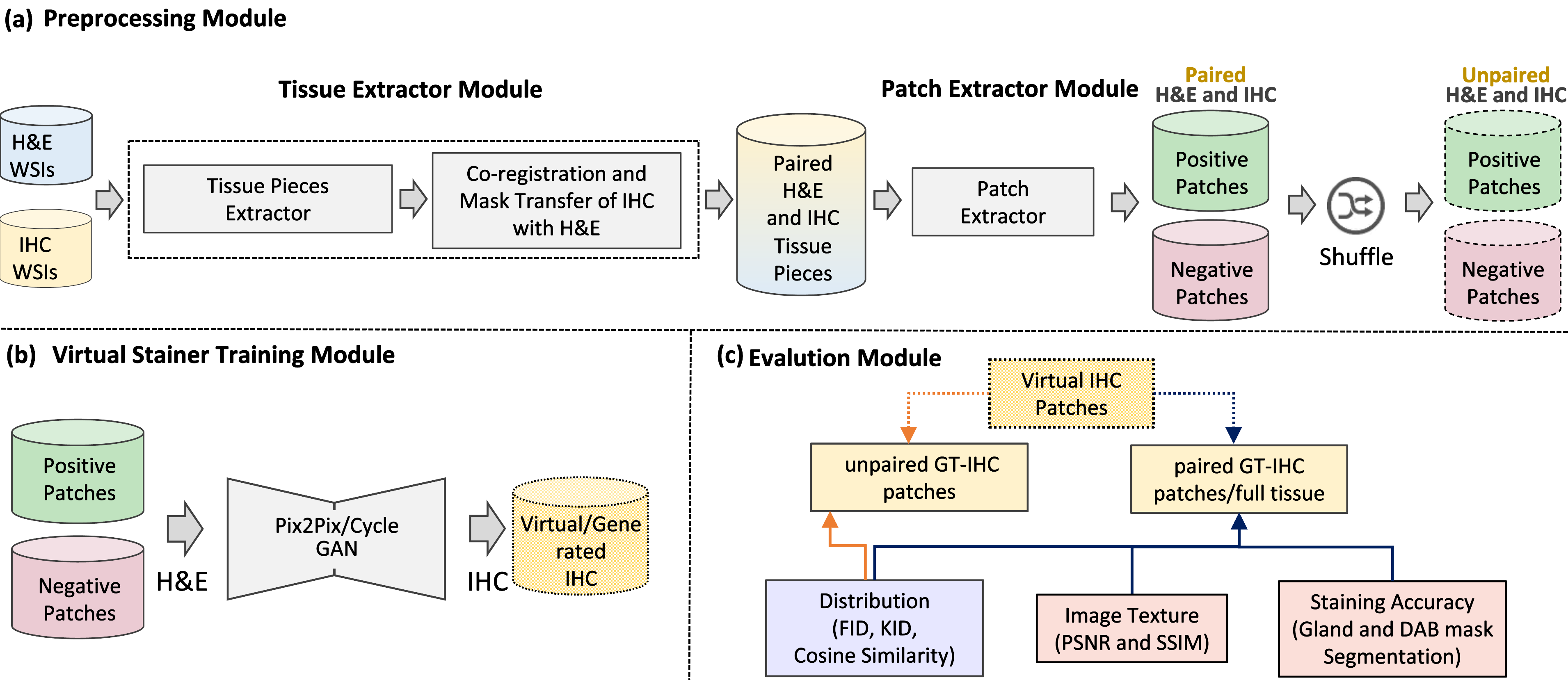}
    \caption{\textbf{Virtual Staining Pipeline:} overview of virtual staining pipeline, comprising: (a) \textbf{Preprocessing module}, including tissue extraction and patch extraction; (b) \textbf{Virtual Staining Training:} model architectures used for paired or unpaired data; and (c) \textbf{Evaluation Module}: A comprehensive automated evaluation system designed to assess the quality and effectiveness of virtually stained images, providing results that are competitive with pathologist assessments.}
    \label{fig:pipeline_full}
\end{figure*}

\begin{figure*}[!htb]
    \centering
    \includegraphics[trim={0.0cm 7.9cm 9.8cm 0.0cm}, clip=true,width=1.0\linewidth]{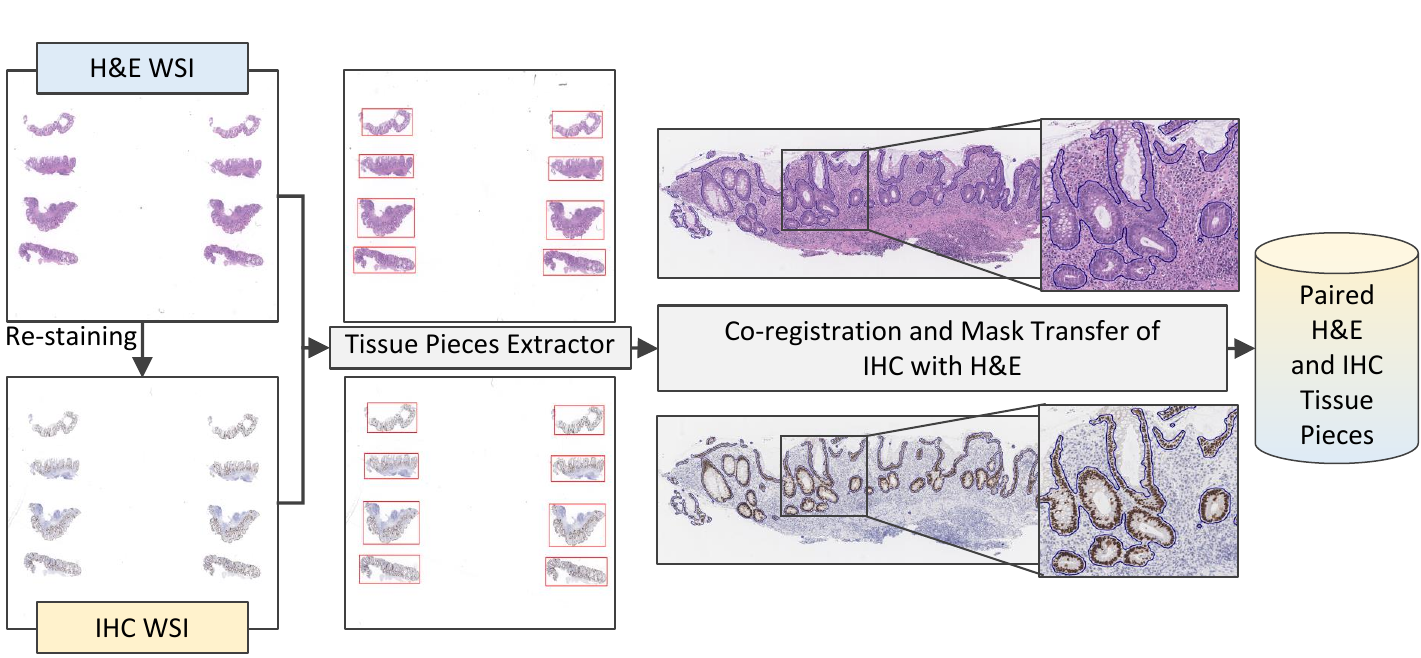}
    \caption{\textbf{Tissue Extraction Module}. The final bounding boxes obtained after initial background removal for both H\&E and IHC stains are shown in the figure above. The right-most images illustrate pixel-wise pairing by transferring the mask—generated through thresholding the IHC DAB channel—to the H\&E image. The mask is manually reviewed to verify that it accurately captures epithelial cells in the H\&E stain, thereby confirming the registration accuracy.}
    \label{fig:TissueExtractionModule}
\end{figure*}

 As shown in Figure \ref{fig:pipeline_full}, this process begins with two whole slide images (WSIs), one stained with H\&E and the other with IHC. Typically, in our dataset, each slide contains 16-24 tissue pieces from the same biopsy block on multiple parallel tissue sections. Since the H\&E and IHC slides are scanned on different scanners, there are no shared reference points to align them directly. These images are 100k pixels in each dimension, so manual registration would be resource-intensive and inefficient.  To address this, we propose an automated tissue registration process consisting of the following steps:
\begin{itemize}
    \item \textit{Downsampling and Grayscale Conversion}: Both H\&E and IHC whole slide images (WSIs) are downsampled by a factor of ten and converted to grayscale.Downsampling is performed to meet the memory limitations of the available CPU, and grayscale conversion is required because the AntsPy library does not support RGB images. We also experimented with alternative color space transformations, such as RGB to HED conversion and using the hematoxylin channel. However, these approaches did not yield better registration results, so we ultimately adopted simple grayscale images.
\item \textit{Thresholding for Tissue Masking}: Otsu’s thresholding is first applied to the grayscale H\&E image to distinguish tissue (foreground) from background. To further refine the resulting tissue masks, morphological operations—specifically dilation and erosion—are applied. These help remove small artifacts caused by staining variability and merge adjacent tissue fragments into unified tissue regions. This process produces clean tissue masks for all tissue sections present in the H\&E WSI. 

\item \textit{Bounding Box}: Contours are extracted from the H\&E tissue masks, and bounding boxes are generated around each tissue region using OpenCV. To exclude irrelevant elements such as noise or artifacts, only contours with an area exceeding a predefined threshold are retained. In our experiments, this threshold was set to 15,000 pixels, although it may require adjustment depending on dataset characteristics.
\footnote{These bounding boxes can be used to extract all tissue sections from the WSI. However, since they are generated from a downsampled version of the WSI (as described in Step 1), the coordinates must be scaled accordingly to match the original resolution.} The bounding boxes obtained for H\&E WSIs through this process are shown in Figure~\ref{fig:TissueExtractionModule}. Currently, bounding boxes are only available for the H\&E images; to complete the pairing, corresponding bounding boxes must also be obtained for the IHC images of all tissue sections.

\item \textit{WSI Registration}: Then the grayscale H\&E image is used as the fixed reference for registering the grayscale IHC image using ANTsPy (\textbf{REF}). We employ ANTs’ deformable registration algorithm, which uses mutual information as the similarity metric and a multi-resolution strategy by default. This is a pixel-wise registration that aims to precisely align corresponding locations between the two grayscale images. The result of this step is a transformation field between the two grayscale images. 

\item \textit{Transformation of Bounding Boxes}: We use the inverse transformation field from the previous registration step to map bounding boxes from H\&E coordinates to the corresponding IHC coordinates. This enables precise extraction of paired tissue regions from the original IHC images, ensuring accurate tissue-level correspondence across WSIs. The bounding boxes transferred to the IHC images through this process are illustrated in Figure~\ref{fig:TissueExtractionModule}.
\end{itemize}

The intermediate results and final bounding boxes from the tissue extraction module are shown in Supplementary Figure~\ref{fig:TissueExtractionModule}. We used the bounding boxes obtained above to extract H\&E and corresponding IHC tissue pieces. Because all steps are performed on downsampled images, the resulting tissue regions for both H\&E and IHC stains are only partially aligned, often exhibiting slight rotations or shifts of several hundred pixels. After evaluating various pixel-level registration algorithms, we determined that the most effective method for achieving precise alignment involves the following steps:
\begin{itemize}
    \item \textit{GrayScale Conversion and Histogram Equalization}. To achieve pixel-level accuracy, we found that performing registration in grayscale after applying histogram equalization produced the most reliable results. This approach was validated across diverse H\&E and IHC stained tissue types. Accordingly, the extracted patches from the previous module are first converted to grayscale, and histogram equalization is then applied to the IHC image to better match the intensity distribution of the H\&E image.
    \item \textit{Multi-Resolution SynRA Normalization}. We employed the ANTsPy library using the SyNRA normalization transform, multi-resolution registration across three levels, and mutual information as the similarity metric to align the grayscale H\&E and IHC images. This step produces a deformation field representing the spatial correspondence between the two images.
    \item \textit{Channel-wise Registeration and Mask Transfer.} The registration transform obtained from the previous step is applied to all channels of the input image to produce the registered images. The accuracy of the registration is verified through manual inspection by two annotators. 
    
    \indent Mask transfer:  an epithelial cell mask is created using the DAB channel of the registered IHC image, obtained by decomposing the RGB image into the HED color space \cite{macenko2009method}. Dilation is applied to eliminate small noise artifacts and minor staining imperfections.  The mask is transferred to the H\&E tissue sections and assessed to determine whether its outline aligns with the contours of epithelial cells, which are readily identifiable in H\&E staining. The segmentation mask and the registered tissue pieces are shown in Figure~\ref{fig:TissueExtractionModule}.
\end{itemize}
Once pixel-level accurate registrations are achieved, patch sampling is done to create the dataset for the virtual staining application, as detailed in the following sections.

\subsection{Patch Extractor Module} \label{suppsec:patchextraction}
Even after background removal, the tissue pieces remain very large—often containing millions of pixels—which makes them unsuitable for direct processing on current GPUs. To enable training of virtual staining models, we extract smaller patches that can efficiently fit into GPU memory. The patch extractor module, illustrated in Figure \ref{fig:pipeline_full}, processes H\&E-stained and associated registered IHC-stained tissue images to prepare paired and unpaired datasets for virtual staining networks. The module is divided into two key components: Area-of-Interest Extractor and Random Patch Extractor.
\begin{figure*}[!htb]
    \centering
      \includegraphics[trim={0.0cm 4.0cm 0cm 0.0cm}, clip=true, width=1.0\linewidth]{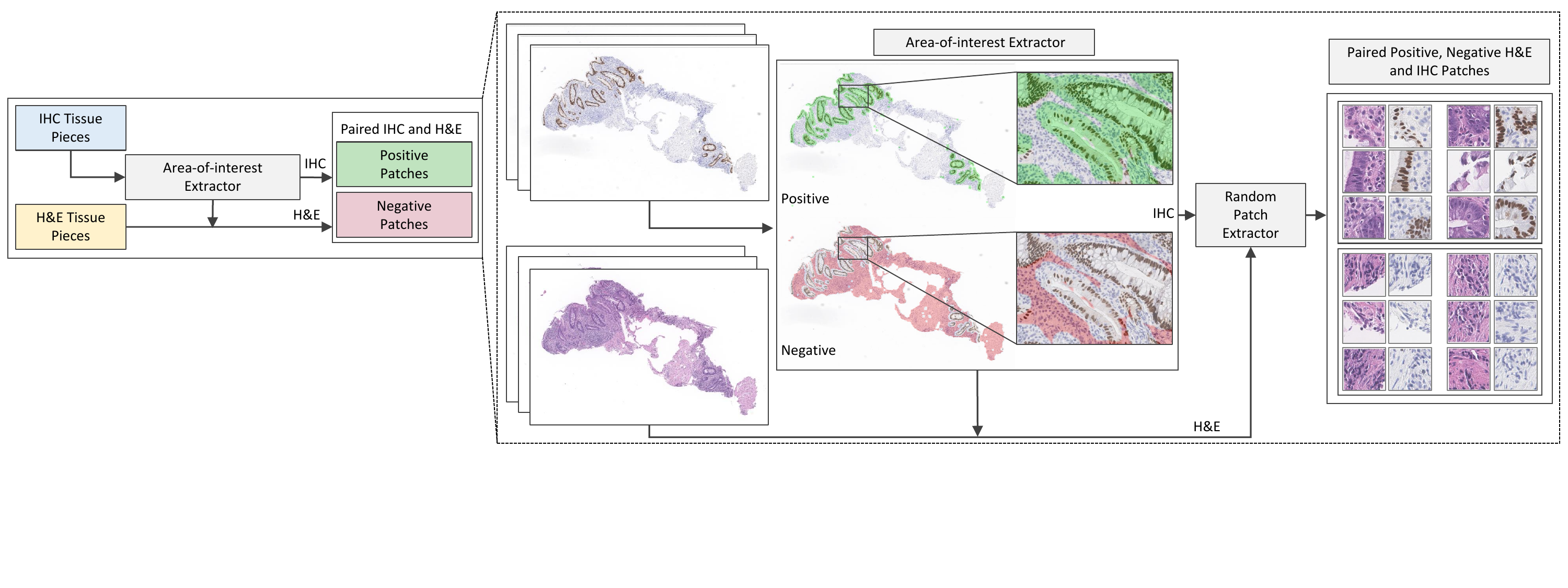}
    \caption{\textbf{Patch Extractor Module}. White tissue and tissue without any brown (positive IHC stain) encompass the majority of the whole tissue. To reduce the oversampling of background patches and negative IHC regions, we propose to do stratified sampling based on the amount of positive and negative regions in the tissue piece.  }
    \label{fig:PatchExtractor}
\end{figure*}
\textit{Area-of-Interest Extractor}. The module identifies two distinct regions in the IHC images: \textbf{Area-of-Interest with positive cells}, where the antibody highlights cells by a brown color; and \textbf{Negative Area-of-Interest}, where no antibody staining is present. The patch extractor operates on the positive and negative areas to balance the dataset. In most IHC stains, the positive regions (highlighted in the brown DAB channel) are significantly smaller than the negative regions. Without balancing, the model may overfit to negative regions and underperform on regions with positive cells — undermining the primary goal of virtual staining, which is to accurately color positive cells.
\begin{itemize}
    \item \textbf{Positive Area-of-Interest Extractor}: The extraction process begins by identifying all regions in the IHC tissue with a DAB mask generated as described in the above section. Around each positive brown pixel, a 32×32 pixels square is defined to capture nearby tissue context. This approach is especially effective where contextual information is critical, such as for brown pixels at the edge of the tissue section and partially or weakly stained areas. When positive pixels are adjacent, their corresponding 32×32 regions intentionally overlap to ensure continuity and comprehensive coverage of brown tissue regions. The resulting mask delineates the mask of the DAB stain within the tissue.
    \item \textbf{Negative Area-of-Interest Extractor}: The negative regions are identified by first isolating the complete tissue area in the IHC image, excluding background whitespace. This is done by converting the image to grayscale and applying a threshold to retain pixels with values above 127. To refine the resulting tissue mask and ensure smooth, continuous boundaries,  erosion and dilation using a 20×20 kernel for 5 iterations are applied. The negative Area-of-Interest is then obtained by subtracting the positive Area-of-Interest from this refined tissue mask, leaving tissue regions that lack marker-specific highlights.

    \item \textbf{Random Patch Extractor}. After separating positive and negative regions in the IHC image, the corresponding areas are mapped onto the paired H\&E image to enable extraction of aligned patches. From each region, 256×256 patches are randomly sampled from both the H\&E and IHC images, ensuring spatial consistency across modalities. This patch size can be adjusted based on GPU memory constraints and the amount of available tissue. The result is a set of paired H\&E-IHC patches. To ensure a balanced dataset, an equal number of positive and negative patches are extracted for each patient.
    
    \item \textbf{Paired and Unpaired Dataset Preparation}: For the paired dataset, the extracted H\&E and IHC patches are directly aligned to maintain correspondence. To create the unpaired dataset, the extracted patches from H\&E and IHC are shuffled independently, removing the alignment between the two modalities. This approach facilitates training both paired and unpaired virtual staining networks.  If the dataset lacks paired tissue samples—i.e., H\&E and IHC stains from the same tissue—the only necessary preprocessing step for the training set is background removal.
\end{itemize}

\section{Virtual Staining Training Models} \label{suppsec:virtualStaining}
Different types of datasets—(a) paired and (b) unpaired—necessitate different model architectures for effective virtual staining. For paired datasets, models such as Pix2Pix family \cite{isola2017image,kataria2024staindiffuser,liu2022bci,li2023bbdm} are commonly used. In contrast, for unpaired datasets, CycleGAN family \cite{zhu2017unpaired,torbunov2023uvcgan,kim2023unpaired,xie2023unpaired,li2023adaptive,dubey2023structural} are more suitable.

\paragraph{Pix2Pix Family.} The Pix2Pix \cite{isola2017image} model architecture consists of two deep learning components: (a)  Generator (G) and (b) Discriminator (D). The Generator follows a U-Net-like architecture, where the input is an image from Domain A, and the model is trained to predict the corresponding image in Domain B, akin to a segmentation task. The Discriminator, on the other hand, is tasked with distinguishing whether the image is real (from Domain B) or generated (by the Generator). This setup creates a min-max optimization problem, where the Generator aims to produce images that are indistinguishable from real images of Domain B, effectively "fooling" the Discriminator. The loss function for this adversarial process is defined as follows: 
\begin{equation*}
    \min_{G} \max_{D} V(D,G) = E_{x \sim p_{data}(x)}[log(D(x))] + E_{z\sim p_z(z)}[log(1-D(G(z))]
\end{equation*}
Additionally, Pix2Pix incorporates an L1 loss as a regularization term to encourage the generated images to closely resemble the target images at a pixel level. Variants of Pix2Pix have been proposed to enhance its performance. For instance, PyramidPix2Pix \cite{liu2022bci} incorporates additional regularization losses and leverages multi-resolution inputs, while VQ-I2I \cite{chen2022eccv} employs a vector-quantized latent space, effectively serving as a latent variant of Pix2Pix. We also consider AdaptiveNCE \cite{li2023adaptive}, which, although conceptually related to CUT \cite{park2020contrastive}—a CycleGAN variant—assumes aligned datasets for training and is therefore more appropriately categorized within the Pix2Pix family.
Overall, Pix2Pix-based methods are well-suited for scenarios where abundant pixel-level paired data between domains is available. However, such datasets are often difficult to obtain in medical imaging, which limits the applicability of Pix2Pix models in medical image translation.

\paragraph{CycleGAN Family.} The CycleGAN \cite{zhu2017unpaired} architecture leverages the cycle consistency property to generalize image-to-image translation tasks for unpaired domains. This property ensures that a generated image in domain B can be mapped back to the original image in domain A and vice versa. The architecture comprises (a) two generators ($G_{A2B}$ and $G_{B2A}$) and (b) two discriminators ($D_A$ and $D_B$), interconnected in a cyclic manner. The losses used to train the CycleGAN model are:

\begin{align*}
    L_{GAN}(G_{A2B},D_B,X,Y) &= E_{y \sim p_{data}(y)}[log(D_B(y))] + E_{x\sim p_data(x)}[log(1-D_B(G_{A2B}(x))] \\
    L_{GAN}(G_{B2A},D_A,Y,X) &= E_{x \sim p_{data}(x)}[log(D_A(x))] + E_{y\sim p_data(y)}[log(1-D_A(G_{B2A}(y))] \\
    L_{cyc}(G_{A2B},G_{B2A}) &= E_{x \sim p_{data}(x)}[|| G_{B2A}(G_{A2B}(x))-x||_1] + E_{y \sim p_{data}(y)}[|| G_{A2B}(G_{B2A}(y))-y||_1]\\
    L_{final} &= L_{GAN}(G_{A2B},D_B,X,Y) + L_{GAN}(G_{B2A},D_A,Y,X) + \lambda L_{cyc}(G_{A2B},G_{B2A})
\end{align*}
In the above equations, $L_{GAN}$ represents the GAN losses for conditional generation (e.g., generating domain A conditioned on input image B, and vice versa). $L_{cyc}$ enforces the cycle consistency that defines the cyclic nature of the CycleGAN architecture. $L_{final}$ is the final loss used to train the model, with $\lambda$ being the hyperparameter. While this model is highly generalizable, it has certain limitations.

Due to its cyclic structure, the model has been shown to embed information from the input domain into high-frequency textures of the generated images \cite{chu2017cyclegan,wu2024stegogan}, leading to reduced generalizability and the potential for hallucinations. Additionally, for some domain translation tasks where one direction is inherently easier than the other (e.g., translating IHC to H\&E is typically easier than H\&E to IHC), the model can develop an internal bias. This imbalance may cause it to converge prematurely to local minima, limiting its performance. Many variants of the CycleGAN architecture have been proposed to address different problems noticed in CycleGAN such as CUT\cite{park2020contrastive}, SC-GAN\cite{dubey2023structural} and others \cite{torbunov2023uvcgan,kim2023unpaired,xie2023unpaired,tang2021attentiongan}.

We focus on establishing an end-to-end pipeline using existing architectures for training and evaluation, designed to assist the broader community in seamlessly applying this pipeline to their specific use cases.  We utilize the original Pix2Pix and CycleGAN architectures from this \href{https://github.com/junyanz/pytorch-CycleGAN-and-pix2pix}{GitHub Link}. For the other models \cite{liu2022bci,li2023adaptive,wu2024stegogan,torbunov2023uvcgan,park2020contrastive,xieunsupervised,li2023bbdm}, we utilize their existing publicly available implementations and retrain all models for the virtual staining task with default parameters. Here is the list of all Image Translation coding repositories used in the paper:
\begin{itemize}
    \item Pix2Pix\cite{isola2017image}: https://github.com/junyanz/pytorch-CycleGAN-and-pix2pix
    \item PyramidPix2Pix\cite{liu2022bci}:  https://github.com/bupt-ai-cz/BCI
    \item AdaptiveNCE \cite{li2023adaptive}:https://github.com/lifangda01/AdaptiveSupervisedPatchNCE
    \item VQ-I2I-Paired \cite{chen2022eccv}:https://github.com/cyj407/VQ-I2I
    \item CycleGAN \cite{zhu2017unpaired}: https://github.com/junyanz/pytorch-CycleGAN-and-pix2pix
    \item CUT \cite{park2020contrastive}: https://github.com/taesungp/contrastive-unpaired-translation
    \item FastCUT \cite{park2020contrastive}: https://github.com/taesungp/contrastive-unpaired-translation
    \item AttentionGAN \cite{tang2021attentiongan}: https://github.com/Ha0Tang/AttentionGAN
    \item Decent GAN \cite{xieunsupervised} : https://github.com/Mid-Push/Decent
    \item QS-GAN \cite{hu2022qs}: https://github.com/sapphire497/query-selected-attention
    \item UNIT\cite{liu2017unsupervised}: https://github.com/mingyuliutw/UNIT
    \item SANTA\cite{xie2023unpaired}: https://github.com/Mid-Push/santa
    \item VQ-I2I \cite{chen2022eccv} : https://github.com/cyj407/VQ-I2I
    \item UVCGAN    \cite{torbunov2023uvcgan} : https://github.com/LS4GAN/uvcgan
    \item StegoGAN  \cite{wu2024stegogan}: https://github.com/sian-wusidi/StegoGAN
    \item UNSB      \cite{kim2023unpaired}: https://github.com/cyclomon/UNSB
\end{itemize}

\begin{table*}[ht]
\centering
\small
\begin{tabular}{l|c|c|p{10cm}}
\textbf{Model} & \textbf{Year} & \textbf{Supervision} & \textbf{Core Idea (Expanded)} \\
\hline
Pix2Pix & 2017 & Paired & Conditional GAN with paired data, combines adversarial loss with L1 reconstruction to enforce pixel-level alignment. \\
\hline
CycleGAN & 2017 & Unpaired & Introduces cycle-consistency loss to translate images between unpaired domains by enforcing forward–backward consistency. \\
\hline
UNIT & 2017 & Unpaired & Unsupervised I2I based on shared latent space assumption using coupled VAEs and GANs; enforces cross-domain latent distribution alignment. \\
\hline
CUT & 2020 & Unpaired & Contrastive unpaired translation using PatchNCE loss, enforcing instance-level correspondence between input and output patches. \\
\hline
FastCUT & 2020 & Unpaired & Simplified version of CUT with one-sided mapping and fewer networks, enabling faster training and inference with reasonable quality. \\
\hline
PyramidPix2Pix & 2021 & Paired & Extends Pix2Pix with pyramid-structured generators and discriminators across multiple scales to better capture global-to-local details. \\
\hline
AttentionGAN & 2021 & Unpaired & Incorporates self-attention modules to adaptively focus on salient regions of the image during translation, improving semantic consistency. \\
\hline
DecentGAN & 2022 & Unpaired & Decomposes the translation task into content and style components, enabling disentangled learning and reducing mode collapse. \\
\hline
QS-GAN & 2022 & Unpaired & Employs query-based attention maps to dynamically match regions across domains, enhancing alignment for complex scene translations. \\
\hline
VQ-I2I & 2022 & Both & Combines vector-quantized variational autoencoders (VQ-VAE) with adversarial learning, mapping images to discrete latent codes for stable translation. \\
\hline
SANTA & 2023 & Unpaired & Self-supervised attention-guided contrastive framework that learns fine-grained correspondence at the patch level without paired data. \\
\hline
UVCGAN & 2023 & Unpaired & Unified contrastive GAN leveraging multi-view consistency and contrastive objectives, particularly effective in medical and remote sensing domains. \\
\hline
AdaptiveNCE & 2023 & Unpaired & Adaptive negative sampling in contrastive objectives to reduce sampling bias, enhancing CUT-like methods on diverse datasets. \\
\hline
StegoGAN & 2024 & Unpaired & Uses steganography-inspired hidden signal embedding for self-supervised guidance, improving representation learning during translation. \\
\hline

UNSB & 2024 & Unpaired & Unified score-based diffusion and GAN framework that integrates diffusion sampling with adversarial objectives for high-quality unsupervised translation. \\

\end{tabular}
\caption{Image-to-Image (I2I) translation models by supervision type and core idea.}
\label{tab:i2i_comparison}
\end{table*}

\section{Evaluation Metrics Details} \label{suppsec:metrics}

\subsection{Distribution Metrics}\label{suppsec:metrics:distribution}
Let $X$ represent the latent space representation of real IHC dataset and $Y$ represent the latent space representation of the virtual generated IHC data, obtained from a pretrained InceptionNet encoder. The distribution metrics are calculation using the following formulations:
\begin{itemize}
    \item \textbf{Frechet Inception Distance(FID).}  Assuming a gaussian distribution of features in the latent space mean and covariance matrics of $X$ and $Y$ are estimated as $\mu_X$, $\mu_Y$ and  $\sum_X$ $\sum_Y$, respectively. Then FID is:
    \begin{equation*}
        FID(X,Y) = || \mu_X-\mu_Y||^2_2 + Tr(\sum_X +\sum_Y -2(\sum_X\sum_Y)^{\frac{1}{2}})
    \end{equation*}
    \item \textbf{Kernel Inception Distance(KID).} KID uses maximum mean discrepancy with a polynomial kernel as distance, which is calculated using:
    \begin{equation*}
        KID(X,Y) = \mathbb{E}[ k(x,x')] + \mathbb{E}[ k(y,y')] -2 \mathbb{E}[ k(x,y)]
    \end{equation*}
    where $k(u,v)=(\frac{1}{d}u^Tv+1)^3$ and $d$ is feature dimension.
    \item \textbf{Precision and Recall.} Precision and recall quantify the fidelity and diversity of generated samples with respect to the real data distribution \cite{kynkaanniemi2019improved}. Let $\mathcal{M}_r$ and $\mathcal{M}_g$ denote the manifolds of real and generated samples in feature space, approximated using k-nearest neighbors:
    \begin{equation*}
        \text{Precision} = \frac{ \{y \in Y | y\in \mathcal{M}_r\} }{|Y|}
    \end{equation*}
     \begin{equation*}
        \text{Recall} = \frac{ \{x \in X | x\in \mathcal{M}_g\} }{|X|}
    \end{equation*}
\end{itemize}

\subsection{Texture Metrics Equations} \label{suppsec:metrics:texture}
Let $\mathbb{I}_{m \times n}$ denote the real ground truth IHC image, where $m\times n$ is the image resolution. Let $\hat{\mathbb{I}}$ represent the predicted (virtually generated) IHC image produced by the virtual staining model. The texture similarity metrics—PSNR, SSIM, and MSE—are then computed using the following equations:
\begin{align*}
    MSE(\mathbb{I},\hat{\mathbb{I}}) &= \frac{1}{m\times n} \sum_m \sum_n ||\mathbb{I}(i,j)-\hat{\mathbb{I}}(i,j)||^2_2\\
    PSNR(\mathbb{I},\hat{\mathbb{I}}) &= 10. log_{10}\frac{255^2}{MSE} \\
    SSIM(\mathbb{I},\hat{\mathbb{I}}) &= \frac{(2\mu_\mathbb{I} \mu_{\hat{\mathbb{I}}} + c_1)(2\sigma_{\mathbb{I}\hat{\mathbb{I}}}+c_2)}{(\mu_\mathbb{I}^2+\mu^2_{\hat{\mathbb{I}}}+c_1)(\sigma^2_{\mathbb{I}} +\sigma^2_{\hat{\mathbb{I}}}+c_2)}
\end{align*}
where $\mu_\mathbb{I}, \mu_{\hat{\mathbb{I}}}$ are sample means of $\mathbb{I}$ and $\hat{\mathbb{I}}$, $\sigma_\mathbb{I}^2, \sigma^2_{\hat{\mathbb{I}}}$ are sample variance of $\mathbb{I}$ and $\hat{\mathbb{I}}$ and $\sigma_{\mathbb{I},\hat{\mathbb{I}}}$ is the sample covariance of $\mathbb{I}$ and $\hat{\mathbb{I}}$. $c_1$ and $c_2$ are constants added to stabilize the denominator.

\subsection{Segmentation Model Training for Segmentation-Based Metrics} \label{suppsec:metrics:modelsegmentation}
Manually annotating IHC images for segmentation would defeat the purpose of automated evaluation. To train the segmentation model, we use thresholding techniques inspired by \textit{Kataria et al.} \cite{kataria2023automating}. Specifically, we threshold the DAB channel and apply morphological operations to create noisy segmentation masks. The model is then trained on these noisy masks, and its predictions are manually reviewed to ensure they accurately match the expected segmentation outcomes.

\subsection{Segmentation Metrics}\label{suppsec:metrics:segmentation}
Assuming $P$ is the predicted segmentation(DAB or model) mask on the generated image and $GT$ is the ground truth segmentation mask on real IHC image the segmentation metrics are calculated as using the following equations:
\begin{equation*}
\centering
Dice = \frac{2*P*GT}{P+GT}
\end{equation*}  
\begin{equation*}
\centering
Jaccard = \frac{P \cap GT }{P \cup GT}
\end{equation*}  
\begin{equation*}
    HD (\text{Hausdorff Distance}) = \text{max}(h(\mathbb{I},\mathbb{\hat{I}}),h(\mathbb{\hat{I}},\mathbb{{I}}))
\end{equation*}
where,
\begin{equation*}
    h(\mathbb{I},\mathbb{\hat{I}}) = \text{sup}_{a\in \mathbb{I}} \text{inf}_{b\in \mathbb{\hat{I}}} d(a,b)
\end{equation*}
\begin{equation*}
    TPR(\text{True Positive Rate}) = \frac{\text{True Positives}}{(\text{True Positives}+\text{False Negatives})}
\end{equation*}

\begin{equation*}
    FNR(\text{False Negative Rate}) = \frac{\text{False Negatives}}{(\text{True Positives}+\text{False Negatives})}
\end{equation*}

\subsection{Code Repositories Used}\label{ssup:code}

For Image Metrics, we used the following Repositories:
\begin{itemize}
    \item FID: https://github.com/mseitzer/pytorch-fid
    \item KID, Distribution Precision and Recall: https://github.com/photosynthesis-team/piq
    \item For calculating PSNR, SSIM and MSE we used https://scikit-image.org/.
    \item For segmentation metrics, we used dice, iou, hausdorff distance, true positive rate and true negative rate from metric in the medpy library(https://loli.github.io/medpy/).  
\end{itemize}

\newpage

\end{document}